\documentclass[11pt]{article}
\usepackage{graphicx}
\usepackage{amsmath, amssymb, enumerate, multirow, multicol}
\usepackage{epstopdf}
\usepackage[T1]{fontenc}
\usepackage{lmodern}
\usepackage{fullpage}
\usepackage{setspace}
\usepackage[usenames,dvipsnames]{xcolor}
\usepackage[colorlinks=true, linkcolor=Blue, urlcolor=Blue, citecolor=Blue, breaklinks=true]{hyperref}
\usepackage{rotating}
\usepackage{bbm}
\usepackage{soul}
\usepackage{comment, verbatim}
\usepackage{pgf, pgffor}
\usepackage{pdflscape}
\usepackage[letterpaper, margin=1in]{geometry}
\usepackage{caption, subcaption}
\usepackage{adjustbox}
\usepackage{booktabs}
\usepackage{dcolumn}
\usepackage{longtable}
\usepackage{tabularx}
\usepackage{threeparttable}
\usepackage{float}
\usepackage{listings}
\usepackage{authblk}
\usepackage[english]{babel}

\usepackage{threeparttable}
\usepackage{changepage}
\usepackage{array}
\usepackage{dashrule}
\usepackage[bottom]{footmisc}
\usepackage{tikz}
\usetikzlibrary{shapes.geometric, arrows, positioning}
\usepackage{indentfirst}

\usepackage[
  backend=biber,
  style=numeric,
  sorting=none,
  citestyle=numeric-comp
]{biblatex}

\usepackage{csquotes}
\defbibenvironment{bibliography}
  {\list
     {\printfield{labelprefix}%
      \printfield{labelnumber}.}
     {\setlength{\leftmargin}{\bibhang}%
      \setlength{\itemindent}{-\leftmargin}%
      \setlength{\itemsep}{\bibitemsep}%
      \setlength{\parsep}{\bibparsep}}}
  {\endlist}
  {\item}

\title{LLMs Model Non-WEIRD Populations: Experiments with Synthetic 
Cultural Agents}

\author[1]{Augusto Gonz\'alez Bonorino}
\author[2,3]{C. M\'onica Capra\thanks{We thank Matthew Feng for research support. We appreciate feedback and comments from Robert Klitgaard, Ernan Haruvy, Daniel Houser, and the late Gary Charness, and participants at the Barcelona Summer School of Economics, the Economic Science Association, the IAREP seminar series, and the Caltech's Frontier Choice Process Mini-Conference. This research was supported by the Blais Challenge Grant at Claremont Graduate University.
\\ Please send correspondence to Prof. M\'onica Capra. Address: 170 E 10th St, Claremont, CA, 91711 USA. E-mail: monica.capra@cgu.edu.
}} 
\author[4]{Emilio Pantoja} 

\affil[1]{Department of Economics, Pomona College}
\affil[2]{Department of Economic Sciences, Claremont Graduate University}
\affil[3]{Center for the Philosophy of Freedom, University of Arizona}
\affil[4]{Department of Economics and Computer Science, Pitzer College}

\begin{document}

\date{}

\maketitle
\thispagestyle{empty}
\begin{abstract}

Despite its importance, studying economic behavior across diverse, non-WEIRD (Western, Educated, Industrialized, Rich, and Democratic) populations presents significant challenges. We address this issue by introducing a novel methodology that uses Large Language Models (LLMs) to create synthetic cultural agents (SCAs) representing these populations. We subject these SCAs to classic behavioral experiments, including the dictator and ultimatum games. Our results demonstrate substantial cross-cultural variability in experimental behavior. Notably, for populations with available data, SCAs' behaviors qualitatively resemble those of real human subjects. For unstudied populations, our method can generate novel, testable hypotheses about economic behavior. By integrating AI into experimental economics, this approach offers an effective and ethical method to pilot experiments and refine protocols for hard-to-reach populations. Our study provides a new tool for cross-cultural economic studies and demonstrates how LLMs can help experimental behavioral research.

\vspace{0.25cm}
\\

\end{abstract}


\newpage
\pagenumbering{arabic}
\setstretch{1.5}

\section{Introduction}

Large language models (LLMs) have emerged as powerful tools for generating human-like text, but their potential for advancing behavioral research remains largely untapped. In this study, we present a novel methodology that leverages LLMs to create synthetic agents representing non-WEIRD (Western, Educated, Industrialized, Rich, and Democratic) populations. By subjecting these agents to established economic experiments, we demonstrate a promising approach for conducting preliminary investigations and generating hypotheses about hard-to-reach populations in an ethical and efficient manner.

The disproportionate representation of WEIRD populations in behavioral research has long been a concern in the social sciences \cite{henrich2020weirdest}. These populations comprise only about 13\% of the global population yet dominate research studies, raising questions about the generalizability of findings to broader human populations. Small-scale societies, in particular, offer valuable insights into human behavior and decision-making, as they more closely represent our evolutionary past \cite{Henrich.weirdest?}. However, studying these populations presents significant challenges, including geographical remoteness, ethical considerations, and methodological inconsistencies across studies \cite{Henrich.2005}.

Our approach presents a new methodology to start addressing these challenges by creating Synthetic Cultural Agents (SCAs), LLM-based models that represent specific cultural profiles. Using a combination of web scraping, LLMs, and retrieval-augmented generation (RAG) prompting, we construct cultural profiles for six small-scale societies: the Hadza, the Machiguenga, the Tsiman\'e, the Ach\'e, the Orma, and the Yanomami. We then use these profiles to instantiate LLM agents and subject them to three classic economic experiments: the dictator game \cite{forsythe1994fairness}, the ultimatum game \cite{guth1982experimental}, and the endowment effect \cite{Kahneman1991}.

Our results reveal substantial cross-cultural variability in experimental behavior, an absence of purely self-interested (\textit{homo economicus}) behavior across all SCAs, and qualitative resemblances to real human populations where data are available. These findings not only generally align with previous research on small-scale societies as described in Henrich, et al. \cite{Henrich.2005} but also extend our understanding to previously unstudied groups.

The alignment of our SCA-generated results with existing human studies validates the potential of LLMs in behavioral research. We build upon the pioneering work of Horton \parencite{horton}, Hewitt et al. \cite{hewitt2024llms}, and Mei et al.\cite{mei2024turing}, who demonstrated that LLMs like ChatGPT can produce results qualitatively similar to human subjects in experiments and can capture human behavioral norms. Our study, however, advances these previous works by deliberately generating cultural profiles for the target populations of interest and incorporating them into the LLM's prompt for economic experiments. That is, we utilize the emerging capability of ``in-context learning'' \cite{dong2024surveyincontextlearning}, which allows us to construct synthetic agents that better reflect the cultural norms, values, and behaviors of specific populations. This method enhances the LLM's ability to represent culturally-specific behaviors in experimental settings.

This study advances the application of artificial intelligence in experimental economics and cultural anthropology, making several key contributions: 
(1) We introduce a novel methodology for creating synthetic cultural agents (SCAs) using large language models (LLMs), providing researchers with a new tool for generating hypotheses and piloting experimental protocols, particularly for hard-to-reach populations. (2) We demonstrate that these SCAs can generate responses aligned qualitatively with key behavioral patterns observed in human subjects from the same societies, while also applying this technique to previously unstudied populations. This dual approach validates the method's potential for preliminary research and illustrates how SCAs can anticipate probable behaviors and inform future research design, even where no prior studies exist.
(3) We present a multimodal platform for refining experimental protocols, exemplified through an implementation of the endowment effect experiment.

Importantly, this work is not intended to replace human participant research, but rather to complement. Researchers can use SCAs to conduct preliminary investigations, refining their hypotheses and experimental designs before undertaking resource-intensive field studies. While our approach shows promise, we acknowledge its limitations, including potential biases inherent in LLMs and the need for careful validation against human data. This study contributes to the ongoing evolution of methodologies in behavioral sciences, potentially expanding our ability to study diverse human populations in a more ethical and efficient manner.

\section{Methods}

The key design principles underlying our methodology are customization and replicability. Specifically, our framework enables researchers to build upon the initial architecture by customizing it to their needs while ensuring that the final model behavior is qualitatively consistent with human behavior reported in the literature. In this paper, we apply the methodology to create cultural profiles of various small-scale societies, instantiate language models instructed to behave as if they were members of the society described by the profile, and subject these synthetic agents to a series of economic games. Through the proposed methods we demonstrate that LLM agents can be used for piloting behavioral studies at lower cost, improving experimental instructions, and studying new and hard-to-reach populations.

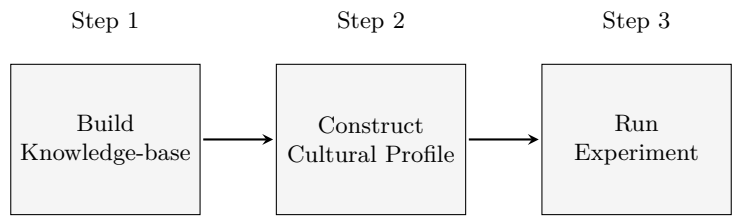
\begin{figure}[!ht]
\centering
\begin{tikzpicture}[
    node font=\footnotesize,
    square/.style={
        rectangle,
        draw,
        minimum width=2.5cm,
        minimum height=2cm,
        text centered,
        text width=2.3cm,
        fill=gray!8
    },
    arrow/.style={
        ->,
        >=stealth,
        thick,
        shorten >=1pt,
        shorten <=1pt
    },
    label/.style={
        text width=2.5cm,
        align=center
    }
]

\node[square] (A) {Build \\Knowledge-base};
\node[square, right=of A] (B) {Construct \\Cultural Profile};
\node[square, right=of B] (C) {Run \\Experiment};

\node[label, above=0.3cm of A] {Step 1};
\node[label, above=0.3cm of B] {Step 2};
\node[label, above=0.3cm of C] {Step 3};

\draw[arrow] (A) -- (B);
\draw[arrow] (B) -- (C);

\end{tikzpicture}
\caption{Framework for Constructing Experiments with Synthetic Cultural Agents}\label{FigFrame}
\end{figure}

Our methodology for creating and experimenting with Synthetic Cultural Agents (SCAs) consists of three key steps: (1) Building a Knowledge Base, (2) Constructing a Cultural Profile, and (3) Running Experiments. This process, illustrated in Figure \ref{FigFrame}, allows for the creation of SCAs that can represent specific small-scale societies in experimental settings and participate in economic decision-making tasks.

We employed this methodology to create SCAs representing six small-scale societies: the Hadza, the Machiguenga, the Tsiman\'e, the Aché, the Orma, and the Yanomami. These societies were chosen for their diversity in economic systems, social structures, and geographical locations, as well as the varying availability of existing experimental data, allowing us to both validate our approach and extend insights to previously unstudied groups.

In constructing the cultural profiles, we considered key socio-cultural-economic factors including lifestyle, cultural practices, economic systems, political ideologies, social organization, kinship structures, and core values \cite{malinowski1922, geertz1973}. These comprehensive profiles inform the behavior of our SCAs, which are then prompted to participate in established economic experiments such as the dictator game, ultimatum game, and tests for the endowment effect.

\subsection{Building the Knowledge Base and Cultural Profile}

To construct accurate and comprehensive cultural profiles, we developed and tested three methodologies: Direct Prompting, Self Ask with Search \cite{press2023measuring}, and Search + Retrieval Augmented Generation (RAG) \cite{lewis2021retrievalaugmented}. After careful evaluation, we determined that the Search + RAG method provided the most reliable and transparent results. Here, we focus on this primary method, with details on the other approaches available in Appendix \ref{App:BuildProfiles} and a discussion about the advantages of RAG over alternatives in Appendix \ref{App:RAGvsFT}.

\subsubsection{Search + RAG Methodology}

Search + RAG combines web scraping, information retrieval models, and large language models to create accurate and up-to-date cultural profiles. This approach addresses limitations of previous methods by enhancing transparency, reducing hallucinations, and allowing for the incorporation of more current information. The process consists of three main components: Search, Retrieval, and Generation.

For building the knowledge base of a tribe, we start with the search query \textit{``What characterizes the (name\_of\_tribe) tribe?''} The search task runs a Google search and returns the URL links of the top $k$ results for the query. The links are fed to a function that scrapes and parses the textual information in each source to store as context. This step is done asynchronously to improve runtime performance. The retrieval task is achieved by splitting the scraped documents into chunks of a fixed length, indexing them in a vector store, and instantiating a retrieval model that ranks chunks based on a similarity measure (e.g., dot product). We implement each step (search, fetch, retrieve, and rank) manually, thus allowing for full customization of the RAG architecture. The retrieved indexed chunks are injected into the profile-building prompt as context for building the cultural profile.

We can control the chunking method, embedding model, choice of vector store, and retriever's parameters. For our experiment we use 2000 words chunk size with 200 words overlap, the BGE open-source embeddings model \cite{chen2024bge}, the open-source Chroma DB vector store to manage our text embeddings and perform similarity searches, with $k=10$. The RAG chains these two components (i.e., Search and RAG) together to enhance the original prompt used to generate a cultural profile with domain-specific and relevant context. That is, the LLM is \textit{instructed to pay attention} to the retrieved information and combines it with the user prompt to generate an answer. A more detailed explanation of the methodology can be found in Appendix \ref{App:SearchRAG} and our \href{https://github.com/Bonorinoa/EconLLM_Lab/tree/main/Projects/Completed/nonWEIRD_Societies/nonWEIRD%20-%20Results}{GitHub repository}.\footnote{https://tinyurl.com/mr4yx3tj}

The Search + RAG methodology offers significant advantages over traditional approaches to creating cultural profiles. By grounding the generation process in retrieved, up-to-date information from web searches, we substantially reduce the risk of hallucinations—factually incorrect statements that can emerge from language models relying solely on pre-trained data. This approach increases the likelihood that our cultural profiles reflect the most current online information about each tribe, potentially capturing recent changes in their socio-economic conditions or cultural practices. Furthermore, the transparency of our method allows for clear traceability of information sources (i.e., url links scraped), as illustrated in Table \ref{tab:RAGSources}, enhancing the credibility and verifiability of our generated profiles. The modular nature of Search + RAG also provides a high degree of customizability, enabling researchers to easily incorporate offline or proprietary data sources, thus adapting the methodology to various research contexts and data availability scenarios. 

The combination of reduced hallucinations, currency of information, transparency, and customizability makes Search + RAG a robust and flexible tool for creating accurate and comprehensive cultural profiles, which serve as the foundation for our Synthetic Cultural Agents (SCAs).


\subsection{Creating Synthetic Cultural Agents and Running Experiments}

Once we have generated comprehensive cultural profiles using the Search + RAG methodology, we use these profiles to instantiate Synthetic Cultural Agents (SCAs) capable of participating in economic experiments. This process involves two key steps: (1) instantiating the SCA with the cultural profile, and (2) subjecting the SCA to experimental tasks.

\subsubsection{Selection of Small-scale Societies}

We created profiles for six small-scale societies: the Hadza, the Machiguenga, the Tsiman\'e, the Aché, the Orma, and the Yanomami. These societies were chosen for their diverse economic and social organizations, as well as the varying availability of existing experimental data. Table \ref{tab:tribes} provides an overview of these societies, including their economic structures, sharing norms, and the presence or absence of previous experimental studies involving our three focal games.

\begin{table}[!hb]
\renewcommand{\arraystretch}{1.3} 
\centering
\begin{adjustbox}{max width=\textwidth}
\small
\begin{tabular}{l>{\raggedright\arraybackslash}p{0.40\textwidth}ccc>{\raggedright\arraybackslash}p{0.20\textwidth}}
\toprule
\textbf{SS-Society} & \textbf{Description} & \textbf{Dict. Game} & \textbf{Ult. Game} & \textbf{End. Eff.} \\
\midrule
Hadza & Hunter-gatherers from northern Tanzania with strong sharing ethics. & \checkmark \cite{marlowe2004dictators} & \checkmark \cite{henrich_15} & \checkmark \cite{Apicella2014}\\
Machiguenga  & Subsistence horticulturalists from the Peruvian Amazon showing cooperation primarily at the family level. & n.d. & \checkmark \cite{Henrich.2000, henrich_15} & n.d. \\
Tsiman\'e & Forager-horticulturalists from the Bolivian Amazon with strong sense of economic independence at the level of the nuclear family and extended household. & \checkmark \cite{gurven2008collective} & \checkmark \cite{henrich_15} & n.d. \\
Ach\'e & Hunter-gatherers from Paraguay with strong norms of sharing catch equally and cooperating among households.\cite{Hill.2002} & n.d. & \checkmark \cite{henrich_15} & n.d. \\
Orma  & Pastoral-nomadic community from Kenya that emphasizes communal sharing and cooperation. & \checkmark \cite{ensminger2004market} & \checkmark \cite{ensminger2004market, henrich_15} & n.d. \\
Yanomami  & Forager-horticulturalists from the Venezuelan and Brazilian Amazon that value cooperation and equitable distribution of resources.\cite{britannica_yanomami} & n.d. & n.d.& n.d. \\
\bottomrule
\end{tabular}
\end{adjustbox}
\caption{Economic Organizations, Sharing Norms and Experimental Studies of Six Tribes}
\label{tab:tribes}
\begin{tablenotes}
    \footnotesize
        \item \textbf{Note:} The table shows the economic organization and sharing norms of the six small-scale societies for which we generate a cultural profile for our synthetic agent. The third - fifth columns show whether there is experimental data for the tribe; \checkmark = available data; n.d. = no documented studies.
    \end{tablenotes}
\end{table}

This diverse selection allows us to both validate our approach against existing data and extend insights to previously unstudied groups, such as the Yanomami.

\subsubsection{Instantiating Synthetic Cultural Agents}

To create an SCA, we use the cultural profile as a system prompt for a large language model (in this case, ChatGPT 3.5). The system prompt instructs the model to respond as if it were a representative member of the specific tribe, sharing similar preferences, viewpoints, and decision-making processes. This approach leverages the model's ability to condition its outputs on the provided context, effectively creating a digital agent that embodies the cultural characteristics of the target population. Details on prompt structure are available in the Appendix \ref{App:Prompts} and our \href{https://github.com/Bonorinoa/EconLLM_Lab/tree/main/Projects/Completed/nonWEIRD_Societies/nonWEIRD%20-%20Results}{GitHub repository.}\footnote{https://tinyurl.com/mr4yx3tj}

We chose ChatGPT 3.5 for this task due to its balance of performance and efficiency. Recent studies have shown that Retrieval-Augmented Generation (RAG) can significantly reduce the required size of the language model by leveraging in-context learning \cite{dong2024surveyincontextlearning}, allowing for the use of less powerful models without compromising performance.\footnote{ChatGPT3.5 is comparatively less powerful than more advanced models, but the RAG prompting approach lowers the required out-of-the-box capabilities of the LLM by explicitly defining the text (i.e., profile) to reference and leveraging in-context learning. We tested ChatGPT4 and Claude 3 Opus as more powerful alternatives to ChatGPT3.5 in the experimental decision step and observed comparable results at a lower computational cost and with faster inference times.} 

\subsubsection{Experimental Setup}

We subjected our SCAs to three classic economic experiments: the dictator game \cite{forsythe1994fairness}, the ultimatum game \cite{guth1982experimental}, and the endowment effect \cite{Kahneman1991}. For the game experiments, we designed prompts that clearly explain the decision environment and the agent's choice set. The prompts are included in the Appendix \ref{App:Prompts}.

In the dictator and ultimatum games, we used endowments equivalent to a day's wage, mirroring the approach used in field experiments with real participants \cite{Henrich.2005}. We employed the strategy method, asking the SCAs to respond with ``yes'' or ``no'' to a series of contingent splits ranging from 0\% to 100\% of the endowment in 10 percentage point increments.

To capture within-subject variation and increase the robustness of our results, we repeated each experimental task 100 times for each SCA. This approach allows us to leverage the inherent stochasticity of the language model while ensuring consistent patterns of behavior across trials.

In addition to the experimental decisions, we prompted each SCA to provide a step-by-step explanation of their reasoning process. This additional data allows us to verify the agent's understanding of the task and gain insights into the cultural factors influencing their decisions. This is akin to Ross et al \cite{ross2024llmeconomicusmappingbehavioral} ``competence tests.'' By prompting the reasoning process, we are able to verify that ChatGPT3.5 can understand endowment splits and perform budget calculations accurately. Responses for each SCA and for each game can be found in our \href{https://github.com/Bonorinoa/EconLLM_Lab/tree/main/Projects/Completed/nonWEIRD_Societies/nonWEIRD%20-%20Results}{GitHub repository}.\footnote{https://tinyurl.com/mr4yx3tj}

For all experiments, we used a fixed temperature of 1, which is the default for API usage of ChatGPT. This setting provides a balance between response randomness and coherence in the model's outputs by influencing the range of words to consider when sampling the next token to generate. 

Finally, for the endowment effect, we took a different approach. As implemented in the field with tribesmen, the endowment effect has a visual component. Indeed, in one of their experiments, Apicella et al. \cite{Apicella2014} placed endowed/exchange items on the ground in front of the Hadza for them to see and choose. In line with this, when developing the endowment effect experiment with SCAs, we employed a two-step multimodal design. First, we generated two items as objects via text-based chat. Next, we tested for the endowment effect by displaying images of the two items to the chatbot, thereby integrating both textual and visual elements into the experimental framework. See Appendix \ref{App:EEPrompts}. Our interactive platform can be accessed through this \href{https://multimodal-egame.streamlit.app/}{link:https://multimodal-egame.streamlit.app/}, and code and documentation can be found through our \href{https://github.com/Bonorinoa/EconLLM_TribesBot} {GitHub repository}.\footnote{https://tinyurl.com/4khsxdky}

\section{Results}

In our research, we examine the behavior of Synthetic Cultural Agents (SCAs) representing six small-scale societies. Our findings support existing anthropological data and offer new insights into unstudied populations. We present our main results from the Dictator Game and Ultimatum Game, followed by an example of our innovative multimodal approach to the Endowment Effect. Additional supporting analysis can be found in Appendix \ref{App:Results}.

\subsection{Dictator Game}

\begin{figure}[htbp]
\centering
    \includegraphics[width=0.7\textwidth]{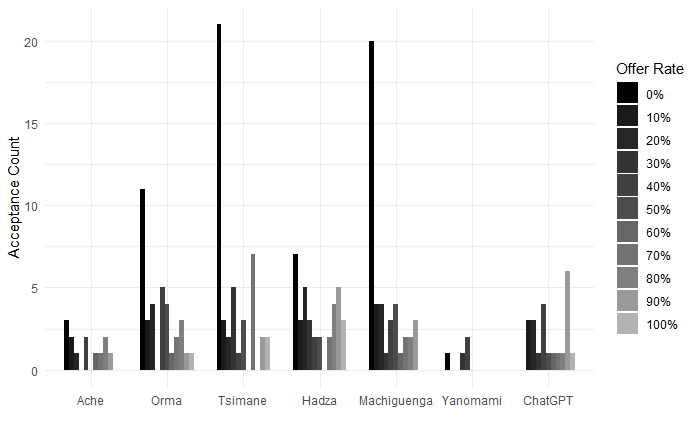}
     \caption{Dictator's Count to Range of Proposed Splits by Tribe and ChatGPT}
     \label{fig:DG_0to100}
        \begin{tablenotes}
            \footnotesize
            \item \textbf{Note:} The figure shows the number of ``yes'' responses by SCAs to proposed splits ranging from 0\% to 100\% in the Dictator Game. Each task (i.e., responding ``yes'' or ``no'') is repeated 100 times. ChatGPT represents the WEIRD benchmark.
        \end{tablenotes}
\end{figure}

The Dictator Game results demonstrate substantial variability in sharing norms across the SCAs, reflecting the diverse cultural backgrounds they represent. Figure \ref{fig:DG_0to100} illustrates the full distribution of dictators' accepted contingent splits across all tribes and ChatGPT. 

A Cochran-Mantel-Haenszel test reveals a significant association between tribal affiliation and acceptance decisions across offer levels (CMH $M^2 = 27.480, p < 0.001$), indicating that cultural factors substantially influence sharing behavior.

\subsection{Ultimatum Game}

The Ultimatum Game results reveal interesting patterns in both proposer and responder behavior across our synthetic agents.

\begin{figure}[!ht]
    \centering
    \includegraphics[width=0.55\linewidth]{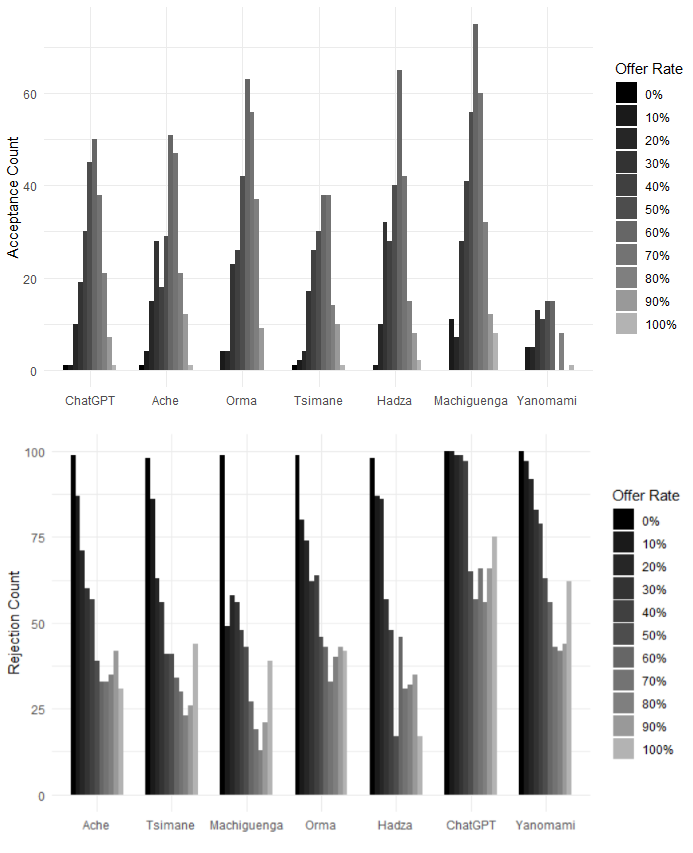}
    \caption{Ultimatum Game Counts to Range of Offers by Tribe and ChatGPT}
    \label{fig:CombinedUG}
    \begin{tablenotes}
            \footnotesize
            \item \textbf{Note:} The top panel of the figure shows the number of Proposer ``yes'' responses by SCAs to contingent offers ranging from 0\% to 100\% in the Ultimatum Game. The bottom panel shows the ``no'' count or rejections to proposed offers by the Responder. Each task (i.e., responding ``yes'' or ``no'') is repeated 100 times. ChatGPT represents the WEIRD benchmark.
        \end{tablenotes}
\end{figure}

Figure \ref{fig:CombinedUG} presents the behaviors in the Ultimatum Game. There are significant variations in both the proposer's (CMH $M^2 = 60.796, p < 0.001$) and the responder's (CMH $M^2 = 27.688, p < 0.001$) behaviors across tribes.

\subsection{Key Findings from Dictator and Ultimatum Games}

Our experiments with Synthetic Cultural Agents (SCAs) reveal significant cross-cultural variability in behavior, reflecting the diverse cultural backgrounds represented by our agents. Notably, SCAs modeled after horticultural societies that primarily cooperate at the family level, such as the Machiguenga or the Tsiman\'e, exhibit more self-interested behavior compared to those representing hunter-gatherer societies like the Aché. This aligns well with anthropological observations of these cultures.

Interestingly, none of the synthetic agents, including the Yanomami (for which we lack real-world experimental data), exhibit purely self-interested behavior consistent with \textit{homo economicus} predictions. This latter finding suggests that our SCAs capture nuanced cultural influences on economic decision-making beyond simple self-interest models.

Our results for the Yanomami, a previously unstudied population in these experimental contexts, provide novel insights. The Yanomami SCA demonstrates behavior closer to \textit{homo economicus} in the Dictator Game and as a proposer in the Ultimatum Game compared to other synthetic agents. However, as a responder in the Ultimatum Game, the Yanomami SCA showed high rejection rates for low offers, indicating a complex interplay of cultural factors in their decision-making processes.

Finally, our use of ChatGPT as a WEIRD benchmark revealed consistently high prosocial behavior. This somewhat diverges from typical behavior observed in WEIRD populations but aligns with previous findings on language models' tendency towards prosociality \cite{mei2024turing, atariwhichhumans}. This latter observation highlights the importance of carefully interpreting ``vanilla'' AI-generated behaviors in the context of human cultural norms.

Overall, the results demonstrate the ability of SCAs to capture cross-cultural variation in economic decision-making by inferring cultural nuances from the generated profiles. The observed patterns align well with existing anthropological data on factors influencing game behavior, such as market integration, community size, and local norms of fairness \cite{gurven2008collective, carpenter2008behavioural, engel2011dictator, Henrich.2005}. Furthermore, for societies that were not previously subjected to experiments, the results provide insights into their plausible behavior. 

\subsection{Endowment Effect: A Multimodal Approach}
\label{sec:multimodal}

We demonstrate our multimodal model with an SCA representing the Ach\'e tribe. This method of implementing the endowment effect simulates Apicella et al's \cite{Apicella2014} experiment, where endowed/exchange items were placed in front of the participants. See Appendices \ref{App:EEPrompts} and \ref{App:EndEFF}. In the example, the Ach\'e SCA rejects the offer to exchange the endowed item (guava fruit) for palm pith. 

As far as we know, no field experimental studies have investigated the endowment effect with the profiled tribes other than the Hadza (see Table \ref{tab:tribes}). Through this example, we demonstrate that our methodology could be used for piloting studies and refining experimental protocols with the Ach\'e and other hard-to-reach populations.

\section{Discussion}

This study introduces Synthetic Cultural Agents (SCAs), a novel methodology leveraging large language models (LLMs) to represent small-scale societies in experimental economics research. By creating SCAs for six diverse small-scale societies and subjecting them to classic economic experiments, we demonstrate the potential of this approach for studying cross-cultural economic behavior, particularly in hard-to-reach populations.

Our results reveal substantial cross-cultural variability which suggests that our SCAs qualitatively capture the cultural nuances influencing economic behavior in these societies. For example, the behavior of SCAs representing horticultural societies like the Machiguenga and Tsiman\'e exhibited more self-interested tendencies compared to those representing hunter-gatherer societies like the Aché, mirroring anthropological observations. However, SCAs consistently demonstrated higher rejection rates in the Ultimatum Game compared to observations from human subjects. This discrepancy may represent a limitation of the SCAs but it could also reflect the ability of our profiling to capture the effect of increased market integration and community size on behavior -- factors known to affect decision-making. Indeed, the more integrated and bigger the tribe is, the closer its behavior is to regular WEIRD. In contrast, highly cooperative, interdependent societies may accept lower offers to prioritize social cohesion over monetary gains. In fact, a key advantage of our methodology (Search + RAG) is that it allows us to observe the evolution of behaviors as new and updated information such as increased market integration is incorporated into the cultural knowledge base that informs our agents' profiles. This dynamic aspect can provide researchers with valuable insights into how evolving cultural and socioeconomic contexts influence behavior.

Our approach offers several potential advantages for experimental economics research. First, it provides a means to efficiently pilot experimental protocols and generate hypotheses about hard-to-reach populations without the ethical concerns and logistical challenges associated with field experiments. Second, the ability to create SCAs for unstudied populations, as demonstrated with the Yanomami, opens new avenues for exploratory research. Third, our multimodal platform more closely captures the implementation of the endowment effect in the field, allowing for more realistic environments for piloting experiments. 

Our current approach to creating profiles of tribal groups and validating SCAs has some limitations. There are constraints related to both the knowledge base we use to build the profiles and how we validate the resulting agents. Our tribal profiles are based on articles, stories, and accounts written mostly by WEIRD people, which represent the majority of available text data -- common sources included Wikipedia, National Geographic, and Nature articles. To create more authentic tribal profiles moving forward, we could use original tribal sources, such as songs and stories collected by ethnographers, or collect primary non-choice data from these tribes to use as input data for our model. This would allow us to build profiles that more accurately represent the knowledge and perspectives of the tribal groups themselves.

Furthermore, our current validation method relies on replicating patterns of behavior observed in the existing literature. However, this approach has its limitations. The ideal way to validate our methodology would be to prospectively profile a hard-to-reach population for which no experimental studies exist and subject the synthetic agent to experiments \textit{before} implementing the study in the field. While this approach would require more resources, it is a natural and logical next step in improving our validation process.

Despite these limitations, the creation of SCA represents a promising step toward expanding the study of cross-cultural economic behavior. By providing a customizable framework for representing diverse populations, SCAs could complement traditional field experiments, offering a useful tool for hypothesis generation, piloting, and refining field experiments \cite{CharnessNBERw31679}. 

Beyond the methodological contribution, by studying synthetic populations that more closely resemble our evolutionary past and observing their development as they grow and establish markets, economists can gain deeper insights into the evolutionary foundations of preferences \cite{capra2011rationality}, \cite{rubin2011evolutionary}. This foundational understanding can inform more accurate representations of human nature in economic models. For instance, the work of Alger and Weibull \cite{alger2020evolutionary}, who incorporate both self-interest and other-regarding preferences in a dynamic evolutionary game framework, and Bester and Güth \cite{bester1998altruism}, who demonstrate the context-dependence of altruism and self-interest, exemplify contributions that could be significantly enriched by insights derived from SCAs representing non-WEIRD populations. SCAs offer a unique opportunity to explore the interplay between evolutionary history, cultural context, and economic behavior, potentially leading to more comprehensive and adaptable economic theories.

\newpage
\noindent \textbf{Declaration of generative AI and AI-assisted technologies in the writing process}

During the preparation of this work the author(s) used Anthropic's Claude in order to edit the manuscript. After using this tool/service, the author(s) reviewed and edited the content as needed and take(s) full responsibility for the content of the publication.
\printbibliography

@article{gurven2008collective,
title={Collective action in action: Prosocial behavior in and out of the laboratory},
author={Gurven, Michael and Winking, Jeffrey},
journal={American Anthropologist},
volume={110},
number={2},
pages={179--190},
year={2008},
publisher={Wiley Online Library}
}

@misc{dong2024surveyincontextlearning,
      title={A Survey on In-context Learning}, 
      author={Qingxiu Dong and Lei Li and Damai Dai and Ce Zheng and Jingyuan Ma and Rui Li and Heming Xia and Jingjing Xu and Zhiyong Wu and Baobao Chang and Xu Sun and Lei Li and Zhifang Sui},
      year={2024},
      eprint={2301.00234},
      archivePrefix={arXiv},
      primaryClass={cs.CL},
      url={https://arxiv.org/abs/2301.00234}, 
}

@misc{hewitt2024llms,
    title={Predicting Results of Social Science Experiments
 Using Large Language Models},
    author={Luke Hewitt and Ashwini Ashokkumar and Isaias Ghezae and Robb Willer},
    year={2024},
    url={https://samim.io/dl/Predicting%20results%20of%20social%20science%20experiments%20using%20large%20language%20models.pdf},
}

@article{mei2024turing,
  title={A Turing Test: Are AI Chatbots Behaviorally Similar to Humans?},
  author={Mei, Qiaozhu and Xie, Yutong and Yuan, Walter and Jackson, Matthew O.},
  journal={arXiv preprint arXiv:2312.00798},
  year={2024}
}

@book{camerer2011behavioral,
  title={Behavioral Game Theory: Experiments in Strategic Interaction},
  author={Camerer, Colin F.},
  year={2011},
  publisher={Princeton University Press}
}

@article{Henrich.2000,
    author = {Henrich, Joseph},
    title = {Does Culture Matter in Economic Behavior? Ultimatum Game Bargaining Among the Machiguenga of the Peruvian Amazon},
    journal = {American Economic Review},
    year = {2000}, 
    doi = {10.1257/aer.90.4.973}
}

@article{Henrich.weirdest?,
  author = {Henrich, J. and Heine, S. and Norenzayan, A.},
  title = {The weirdest people in the world?},
  journal = {Behavioral and Brain Sciences},
  year = {2010},
  volume = {33},
  issue = {2-3},
  pages = {61-83},
  doi = {10.1017/s0140525x0999152x}
}

@article{Henrich.2005,
  author = {Henrich, Joseph and Boyd, Robert and Bowles, Samuel and Camerer, Colin F. and Fehr, Ernst and Gintis, Herbert and McElreath, Richard and Alvard, Michael and Barr, Abigail and Ensminger, Jean and Henrich, Natalie and Hill, Kim and Gil-White, Francisco J. and Gurven, Michael and Marlowe, Frank W. and Patton, John Q. and Tracer, David P.},
  doi = {10.1017/s0140525x05000142}, pages = {795-815},
  title = {"Economic Man" in Cross-Cultural Perspective: Behavioral Experiments in 15 Small-Scale Societies},
  journal = {Behavioral and Brain Sciences},
  year = {2005}
}

@book{henrich2020weirdest,
  author = {Henrich, Joseph},
  title = {The WEIRDest People in the World: How the West Became Psychologically Peculiar and Particularly Prosperous},
  publisher = {Farrar, Straus and Giroux},
  year = {2020},
  address = {New York},
  isbn = {978-0374173227}
}

@article{henrich_15,
  author = {Henrich, Joseph and Boyd, Robert and Bowles, Samuel and Camerer, Colin and Fehr, Ernst and Gintis, Herbert and McElreath, Richard},
  title = {In Search of Homo Economicus: Behavioral Experiments in 15 Small-Scale Societies},
  journal = {American Economic Review},
  volume = {91},
  number = {2},
  year = {2001},
  pages = {73-78},
  doi = {10.1257/aer.91.2.73},
  url = {https://www.aeaweb.org/articles?id=10.1257/aer.91.2.73}
}

@article{horton,
  title = {Large Language Models as Simulated Economic Agents: What Can We Learn from Homo Silicus?},
  author = {John J. Horton},
  year = {2023},
  eprint = {2301.07543},
  archivePrefix = {arXiv},
  primaryClass = {econ.GN}
}

@article{Apicella2014,
  title = {Evolutionary Origins of the Endowment Effect: Evidence from Hunter-Gatherers},
  author = {Apicella, Coren L. and Azevedo, Eduardo M. and Christakis, Nicholas A. and Fowler, James H.},
  journal = {American Economic Review},
  volume = {104},
  number = {6},
  pages = {1793--1805},
  year = {2014},
  publisher = {American Economic Association},
  doi = {10.1257/aer.104.6.1793}
}

@book{malinowski1922,
  title = {Argonauts of the Western Pacific: An Account of Native Enterprise and Adventure in the Archipelagoes of Melanesian New Guinea},
  author = {Malinowski, Bronislaw},
  year = {1922},
  publisher = {George Routledge \& Sons, Ltd.},
  address = {London}
}

@book{geertz1973,
  title = {The Interpretation of Cultures: Selected Essays},
  author = {Geertz, Clifford},
  year = {1973},
  publisher = {Basic Books},
  address = {New York}
}

@article{Kahneman1991,
  title = {Anomalies: The Endowment Effect, Loss Aversion, and Status Quo Bias},
  author = {Kahneman, Daniel and Knetsch, Jack L. and Thaler, Richard H.},
  journal = {Journal of Economic Perspectives},
  volume = {5},
  number = {1},
  pages = {193--206},
  year = {1991},
  publisher = {American Economic Association},
  doi = {10.1257/jep.5.1.193}
}

@article{li2023selfprompting,
  title = {Self-Prompting Large Language Models for Zero-Shot Open-Domain QA},
  author = {Junlong Li and Zhuosheng Zhang and Hai Zhao},
  year = {2023},
  eprint = {2212.08635},
  archivePrefix = {arXiv},
  primaryClass = {cs.CL}
}

@misc{lewis2021retrievalaugmented,
      title={Retrieval-Augmented Generation for Knowledge-Intensive NLP Tasks}, 
      author={Patrick Lewis and Ethan Perez and Aleksandra Piktus and Fabio Petroni and Vladimir Karpukhin and Naman Goyal and Heinrich Küttler and Mike Lewis and Wen-tau Yih and Tim Rocktäschel and Sebastian Riedel and Douwe Kiela},
      year={2021},
      eprint={2005.11401},
      archivePrefix={arXiv},
}

@misc{press2023measuring,
  title = {Measuring and Narrowing the Compositionality Gap in Language Models},
  author = {Ofir Press and Muru Zhang and Sewon Min and Ludwig Schmidt and Noah A. Smith and Mike Lewis},
  year = {2023},
  eprint = {2210.03350},
  archivePrefix = {arXiv},
  primaryClass = {cs.CL}
}

@misc{ross2024llmeconomicusmappingbehavioral,
      title={LLM economicus? Mapping the Behavioral Biases of LLMs via Utility Theory}, 
      author={Jillian Ross and Yoon Kim and Andrew W. Lo},
      year={2024},
      eprint={2408.02784},
      archivePrefix={arXiv},
      primaryClass={cs.CL},
      url={https://arxiv.org/abs/2408.02784}, 
}

@article{wei2023chainofthought,
  title = {Chain-of-Thought Prompting Elicits Reasoning in Large Language Models},
  author = {Jason Wei and Xuezhi Wang and Dale Schuurmans and Maarten Bosma and Brian Ichter and Fei Xia and Ed Chi and Quoc Le and Denny Zhou},
  year = {2023},
  eprint = {2201.11903},
  archivePrefix = {arXiv},
  primaryClass = {cs.CL}
}

@article{yao2023react,
  title = {ReAct: Synergizing Reasoning and Acting in Language Models},
  author = {Shunyu Yao and Jeffrey Zhao and Dian Yu and Nan Du and Izhak Shafran and Karthik Narasimhan and Yuan Cao},
  year = {2023},
  eprint = {2210.03629},
  archivePrefix = {arXiv},
  primaryClass = {cs.CL}
}

@misc{liu2023lost,
  title = {Lost in the Middle: How Language Models Use Long Contexts},
  author = {Nelson F. Liu and Kevin Lin and John Hewitt and Ashwin Paranjape and Michele Bevilacqua and Fabio Petroni and Percy Liang},
  year = {2023},
  eprint = {2307.03172},
  archivePrefix = {arXiv},
  primaryClass = {cs.CL}
}

@article{guth1982experimental,
  title = {An experimental analysis of ultimatum bargaining},
  author = {G{\"u}th, Werner and Schmittberger, Rolf and Schwarze, Bernd},
  journal = {Journal of economic behavior \& organization},
  volume = {3},
  number = {4},
  pages = {367--388},
  year = {1982},
  publisher = {Elsevier}
}

@article{forsythe1994fairness,
  title = {Fairness in simple bargaining experiments},
  author = {Forsythe, Robert and Horowitz, Joel L and Savin, Nathan E and Sefton, Martin},
  journal = {Games and economic behavior},
  volume = {6},
  number = {3},
  pages = {347--369},
  year = {1994},
  publisher = {Elsevier}
}

@misc{chen2024bge,
  title = {BGE M3-Embedding: Multi-Lingual, Multi-Functionality, Multi-Granularity Text Embeddings Through Self-Knowledge Distillation},
  author = {Jianlv Chen and Shitao Xiao and Peitian Zhang and Kun Luo and Defu Lian and Zheng Liu},
  year = {2024},
  eprint = {2402.03216},
  archivePrefix = {arXiv},
  primaryClass = {cs.CL}
}

@article{henrich2006costly,
  title={Costly punishment across human societies},
  author={Henrich, Joseph and McElreath, Richard and Barr, Abigail and Ensminger, Jean and Barrett, Clark and Bolyanatz, Alexander and Cardenas, Juan Camilo and Gurven, Michael and Gwako, Edwins and Henrich, Natalie and Lesorogol, Carolyn and Marlowe, Frank and Tracer, David and Ziker, John},
  journal={Science},
  volume={312},
  number={5781},
  pages={1767--1770},
  year={2006},
  publisher={American Association for the Advancement of Science},
  doi={10.1126/science.1127333}
}

@article{carpenter2008behavioural,
  title={Behavioural Development Economics: Lessons from Field Labs in the Developing World},
  author={Carpenter, Jeffery and Cardenas, Juan Camilo},
  journal={The Journal of Development Studies},
  volume={44},
  number={3},
  pages={311-338},
  year={2008},
  publisher={Taylor \& Francis},
  doi={10.1080/00220380701848327}
}

@article{Hill.2002,
    author = "Hill, Kim",
    title = "Altruistic Cooperation During Foraging by the Ache, and the Evolved Human Predisposition to Cooperate",
    journal = "Human Nature",
  volume={13},
  number={1},
  pages={105-128},
    year = {2002}, doi = "10.1007/s12110-002-1016-3",
}

@incollection{marlowe2004dictators,
  title={Dictators and Ultimatums in an Egalitarian Society of Hunter-Gatherers: The Hadza of Tanzania},
  author={Marlowe, Frank},
  booktitle={Foundations of Human Sociality: Economic Experiments and Ethnographic Evidence from Fifteen Small-Scale Societies},
  editor={Henrich, Joseph and Boyd, Robert and Bowles, Samuel and Camerer, Colin and Fehr, Ernst and Gintis, Herbert},
  pages={168--193},
  year={2004},
  publisher={Oxford University Press},
  address={Oxford}
}

@incollection{ensminger2004market,
  title={Market Integration and Fairness: Evidence from Ultimatum, Dictator, and Public Goods Experiments in East Africa},
  author={Ensminger, Jean},
  booktitle={Foundations of Human Sociality: Economic Experiments and Ethnographic Evidence from Fifteen Small-Scale Societies},
  editor={Henrich, Joseph and Boyd, Robert and Bowles, Samuel and Camerer, Colin and Fehr, Ernst and Gintis, Herbert},
  pages={356--381},
  year={2004},
  publisher={Oxford University Press},
  address={Oxford}
}

@article{engel2011dictator,
  title={Dictator games: a meta study},
  author={Engel, Christoph},
  journal={Experimental Economics},
  volume={14},
  number={4},
  pages={583--610},
  year={2011},
  publisher={Springer},
  doi={10.1007/s10683-011-9283-7}
}

@misc{britannica_yanomami,
  author = {{The Editors of Encyclopaedia Britannica}},
  title = {Yanomami},
  year = {2024},
  url = {https://www.britannica.com/topic/Yanomami},
  note = {Accessed: 2024-07-10},
  howpublished = {\url{https://www.britannica.com/topic/Yanomami}}
}

@article{plott2007exchange,
  title={Exchange Asymmetries Incorrectly Interpreted as Evidence of Endowment Effect Theory and Prospect Theory},
  author={Plott, Charles R. and Zeiler, Kathryn},
  journal={American Economic Review},
  volume={97},
  number={4},
  pages={1449--1466},
  year={2007},
  doi={10.1257/aer.97.4.1449}
}

@article{blurtonjones1984selfish,
  title={A selfish origin for human food sharing: Tolerated theft},
  author={Blurton Jones, N. G.},
  journal={Ethology and Sociobiology},
  volume={5},
  number={1},
  pages={1--3},
  year={1984},
  publisher={Elsevier},
  doi={10.1016/0162-3095(84)90030-X}
}

@incollection{rubin2011evolutionary,
  title={The evolutionary psychology of economics},
  author={Rubin, Paul H and Capra, C Monica},
  booktitle={Applied Evolutionary Psychology},
  editor={Roberts, S Craig},
  year={2011},
  publisher={Oxford University Press},
  address={Oxford}
}

@incollection{capra2011rationality,
  title={Rationality and Utility: Economics and Evolutionary Psychology},
  author={Capra, C. Monica and Rubin, Paul H.},
  booktitle={Evolutionary Psychology in the Business Sciences},
  editor={Saad, Gad},
  year={2011},
  publisher={Springer},
  pages={319-338}
}

@techreport{CharnessNBERw31679,
 title = "Generation Next: Experimentation with AI",
 author = "Charness, Gary and Jabarian, Brian and List, John A",
 institution = "National Bureau of Economic Research",
 type = "Working Paper",
 series = "Working Paper Series",
 number = "31679",
 year = "2023",
 month = sep,
 doi = {10.3386/w31679},
 URL = "http://www.nber.org/papers/w31679",
}

@article{bester1998altruism,
  title={Is altruism evolutionarily stable?},
  author={Bester, Helmut and Güth, Werner},
  journal={Journal of Economic Behavior \& Organization},
  volume={34},
  number={2},
  pages={193--209},
  year={1998},
  publisher={Elsevier}
}

@article{alger2020evolutionary,
  title={Evolutionary models of preference formation},
  author={Alger, Ingela and Weibull, J{\"o}rgen W.},
  journal={Annual Review of Economics},
  volume={12},
  pages={329--354},
  year={2020},
  publisher={Annual Reviews}
}

@misc{atariwhichhumans,
 title={Which Humans?},
 url={osf.io/preprints/psyarxiv/5b26t},
 DOI={10.31234/osf.io/5b26t},
 publisher={PsyArXiv},
 author={Atari, Mohammad and Xue, Mona J and Park, Peter S and Blasi, Damián E and Henrich, Joseph},
 year={2023},
 month=sep
}

\newpage
\thispagestyle{empty}
\begin{center}\textbf{Data Availability
}\end{center}

\begin{enumerate}
    \item \textbf{Repository:} The Python code for creating synthetic cultural agents (SCAs), the interactive platform, and the data generated from SCA simulations are available in our \href{https://github.com/Bonorinoa/EconLLM_Lab/tree/main/Projects/Completed/nonWEIRD_Societies/nonWEIRD%20-%20Results}{GitHub repository: https://tinyurl.com/mr4yx3tj}. This repository is publicly accessible and includes detailed README files for navigation and usage.

    \item \textbf{Data Types:} The repository contains raw SCA output data, processed experimental data, configuration files, and source code for the interactive platform.

    \item \textbf{Large Language Models:} We used GPT-3.5 turbo for creating SCAs, and GPT4, GPT4o, and Claude 3 Opus for creating the profiles.

    \item \textbf{License:} All data and code are freely available for academic use under the MIT License.

    \item \textbf{Supplemental Data:} Tables containing data on the number of observations for each proposed endowment split in the Dictator and Ultimatum Games can be found in the Supplemental Materials (Appendices).

    \item \textbf{Data Sufficiency:} All data supporting the findings of this study are available within the paper and its supplementary materials, ensuring full replicability of the results.

    \item \textbf{Additional Information:} Readers are encouraged to contact the corresponding author (\href{mailto:monica.capra@cguedu}{monica.capra@cguedu}) for any additional data-related queries or specific requests.
\end{enumerate}


\newpage

\setcounter{page}{0}
\pagenumbering{arabic}
\setcounter{page}{1}

\setcounter{section}{0}
\renewcommand*\thesection{Appendix \Alph{section}:}
\renewcommand*\thesubsection{\Alph{section}.\arabic{subsection}}
\setcounter{table}{0}
\renewcommand*\thetable{\Alph{section}.\arabic{table}}
\setcounter{figure}{0}
\renewcommand*\thefigure{\Alph{section}.\arabic{figure}}

\begin{center}
{\bf{\LARGE{For Online Publication}\vspace{0.1in}\\ \large{
LLM Experiments with Synthetic Tribesmen}}}
\section*{Appendix}\label{sec:appendix}
\end{center}

\appendix
\renewcommand{\thetable}{A\arabic{table}}
\renewcommand{\thefigure}{A\arabic{figure}}
\counterwithin{figure}{section}
\counterwithin{table}{section}

\section{Methods} \label{App:BuildProfiles}

\subsection{Build Cultural Profile}
\subsubsection{Direct Prompting} \label{App:Dir}

    Direct Prompting consists of simply prompting the LLM to generate a profile by primarily sampling information from its training data, conditioned on a list of relevant factors (input as a list of keywords to consider). We constructed this prompt in a "parameterized" manner, which helps isolate the main structure of the prompt and thus modularize the code. In this case, the parameters are the tribe's name and keywords to guide the output's structure. 

    \begin{lstlisting}
    def build_profile(prompt: str):
    '''
    Function to initialize and run a GPT model to build a cultural profile from its training data
    params
        prompt (str): prompt to build the profile
    '''
    llm = ChatOpenAI(model_name='gpt-4',
                     temperature=0.5,
                     max_tokens=500)

    system_prompt = "You're a helpful assistant that aids 
                in constructing detailed and comprehensive cultural profiles"

    messages = [
            SystemMessage(content=system_prompt),
            HumanMessage(content=prompt),
        ]

    response = llm.invoke(prompt)

    return response
    
    # prompt inputs for search
    tribe_to_search = "Hazda"
    relevant_factors = ["lifestyle," "average age", "culture", "economic system", "political ideologies", "values", "kinship", Social Organization"]
    
    # Generate persona based on profile and system prompt
    prompt = f"Please construct a profile on the {tribe_to_search}. " \
            + f" The profile must cover the following socio-economic relevant factors {relevant_factors}. Proceed step by step."
    
    tribe_profile1 = build_profile(prompt)
    \end{lstlisting}

    Direct prompting is the methodology most closely aligned with the approaches used by both Horton \cite{horton} and Mei et al. \cite{mei2024turing}. In this scenario, a cultural profile is built solely based on the training data the LLM has access to and a specific temperature chosen before generation. Although important insights can be gained from LLMs prompted to acquire a cultural profile, there are problems with this approach. 
    
    We find that the generations often needed tuning (i.e., further chat to correct output), were unreliable either due to hallucinations or lack of sources it referenced and did not meet the maximum tokens requirement (this caused profiles to be cut short). Nevertheless, it is the fastest approach because it involves no additional steps to curate context. Recall that hallucinations are generations of the model that are coherent and grammatically correct yet factually incorrect, qualities we would like to prevent in our cultural profile.

    LLMs are pretrained language models, which means that the training data they access is static at the time of usage. In other words, an LLM knowledge base is everything on the internet up to a certain date (i.e., the training cutoff). This leaves out the possibility of referencing the most current available information when building the cultural profile.


\subsubsection{LLM Agent with Search}\label{App:Lang}

LangChain's ``SelfAsk with Search'' agent is an LLM model equipped with tools that enable it to automate tasks \cite{li2023selfprompting}. The agent is built into the LangChain Python module \footnote{\href{https://python.langchain.com/v0.1/docs/modules/agents/agent_types/self_ask_with_search/}{SelfAsk with Search Agent}}, thus it is freely and easily accessible. This methodology relies on fetching relevant information autonomously from the web and generating a Cultural Profile (steps 1 and 2) by combining the self-ask prompting method with a Google search tool. The search tool endows an LLM with the capability of browsing the internet given a query \cite{press2023measuring}. The self-ask prompt builds upon Chain-of-Thought \cite{wei2023chainofthought} by applying a Reason-Act \cite{yao2023react} framework. This guides the model's chain of reasoning via a series of automatically constructed follow-up questions constrained by the relevant socio-economic factors specified in the prompt.

 Table \ref{tab:LangChain} shows an excerpt of the Agent's execution chain given the prompt shown in \ref{App:Dir}. Once no more follow-up questions are generated, or a maximum number of iterations is reached, the LLM parses through the intermediate answers and generates a cultural profile. Our \href{https://github.com/Bonorinoa/EconLLM_Lab/tree/main/Projects/Completed/nonWEIRD_Societies/nonWEIRD%20-%20Results}{GitHub repository} includes code along with detailed explanations of our methodology.

    \begin{lstlisting}
    def run_search_agent(prompt_search: str):
    '''
    Function to initialize and run a LangChain SELF_ASK_WITH_SEARCH Agent with access to google serper.
    params:
        prompt (str): Instructions for the agent about what to search.
    '''
    # Initialize the LLM
    llm = ChatOpenAI(model_name='gpt-4',
                     temperature=0.5,
                     max_tokens=500)

    search = GoogleSearchAPIWrapper()
    tools = [
        Tool(
            name="Intermediate Answer",
            func=search.run,
            description="useful for when you need to ask with search",
        )
    ]

    self_ask_with_search = initialize_agent(tools, llm,
                                            agent=AgentType.SELF_ASK_WITH_SEARCH,
                                            verbose=True,
                                            max_iterations=10,
                                            early_stopping_method="generate",
                                            handle_parsing_errors=True)

    result = self_ask_with_search.run(prompt_search)

    return result

    # prompt inputs for search
    tribe_to_search = "Hazda"
    relevant_factors = ["lifestyle," "average age", "culture", "economic system", "political ideologies", "values", "kinship", Social Organization"]
    
    # Generate persona based on profile and system prompt
    prompt_search = f"Please construct a detailed and comprehensive cultura profile on the {tribe_to_search}. " \
            + f" The profile must cover the following socio-economic relevant factors {relevant_factors}, use search to get this information."
    
    tribe_profile2 = run_search_agent(prompt_search=prompt_search)
    \end{lstlisting}

    \setlength{\tabcolsep}{10pt} 
    \begin{table}[!ht]
    \centering \footnotesize
    \begin{tabularx}{\textwidth}{|X|}
      \hline
\begin{lstlisting}
Follow up: What is the lifestyle of the Yanomami?

Intermediate answer: The Yanomami are the largest relatively isolated tribe in South America. They live in the rainforest and mountains in northern Brazil and southern ... Mar 2, 2024 ... Yanomami, South American Indians, speakers of a Xiriana language, who live in the remote forest of the Orinoco River basin in southern ... Aug 9, 2022 ... Today, the Yanomami - who number about 29,000 - say they are at serious risk of losing their lands, culture and traditional way of life. The ... Nov 15, 2018 ... The Yanomami diet, low in fat and salt and high in fiber, consists of such items as plantains, cassavas (a root vegetable), fruit, and meat- [...]

Follow up: What is the culture of the Yanomami?

Intermediate answer: The Yanomami are one of the most numerous, and best-know, forest-dwelling tribes in South America. Their home is in the Amazon rainforest, among the hills...Mar 2, 2024... Yanomami, South American Indians, speakers of a Xiriana language, who live in the remote forest on the Orinoco River basin in southern...Aug 9, 2022 ... Today, the Yanomami - who number about 29,000 - say they are at serious risk of losing their lands, culture and traditional way of life. The ... The Yanomami, also spelled Yanomamo or Yanomama, are a group of approximately 35,000 indigenous people who live in some 200-250 villages in the Amazon [...]

Follow up: What is the economic system of the Yanomami?

Intermediate answer:  [...]
\end{lstlisting}\\
      \hline
    \end{tabularx}
    \caption{Example of a LangChain Agent Execution Chain}
    \label{tab:LangChain}
    \begin{tablenotes}
        \footnotesize
            \item \textbf{Note:} Code snippet depicting the LangChain agent's chain of reasoning. The agent parses the \{relevant\_factors\} such as lifestyle, and culture one at a time. In the model, once no more follow-up questions are generated, or a maximum number of iterations is reached, the LLM parses through the intermediate answers (e.g., ``The Yanomami are the largest...'') to generate a cultural profile.
        \end{tablenotes}
    \end{table}
    \setlength{\tabcolsep}{6pt}

    The profile was of higher quality and did not hallucinate throughout our experiments. But several key limitations emerged. First, the sources referenced were not evident and no immediate way of constraining the sources to visit exist. Second, the profile was often too short and information was tightly compressed. Third, runtime scales badly with number of follow up questions. Fourth, the number of tokens used is significantly higher. Therefore, this architecture provides a more factually consistent generation at the cost of runtime and tokens.

\subsubsection{Search + RAG} \label{App:SearchRAG}

    Enhancing the system prompt with information from the web proved helpful, but its lack of transparency and inefficiency makes using built-in Agents a "black-box" hard to manipulate. This third architecture decouples the search and retrieve task from the LLM, thus providing full control at each step of the methodology. 

    Figure \ref{fig:RAG} depicts this methodology. The systems starts by running a google search based on an input query and returns the url links of the top k results. These links are fed to a function that scrapes and parse the textual information in each source to store as potential context. We run our experiment with the search query "What characterizes the hazda tribe?". 

  \begin{figure}[!ht]
    \centering
    \includegraphics[width=\textwidth]{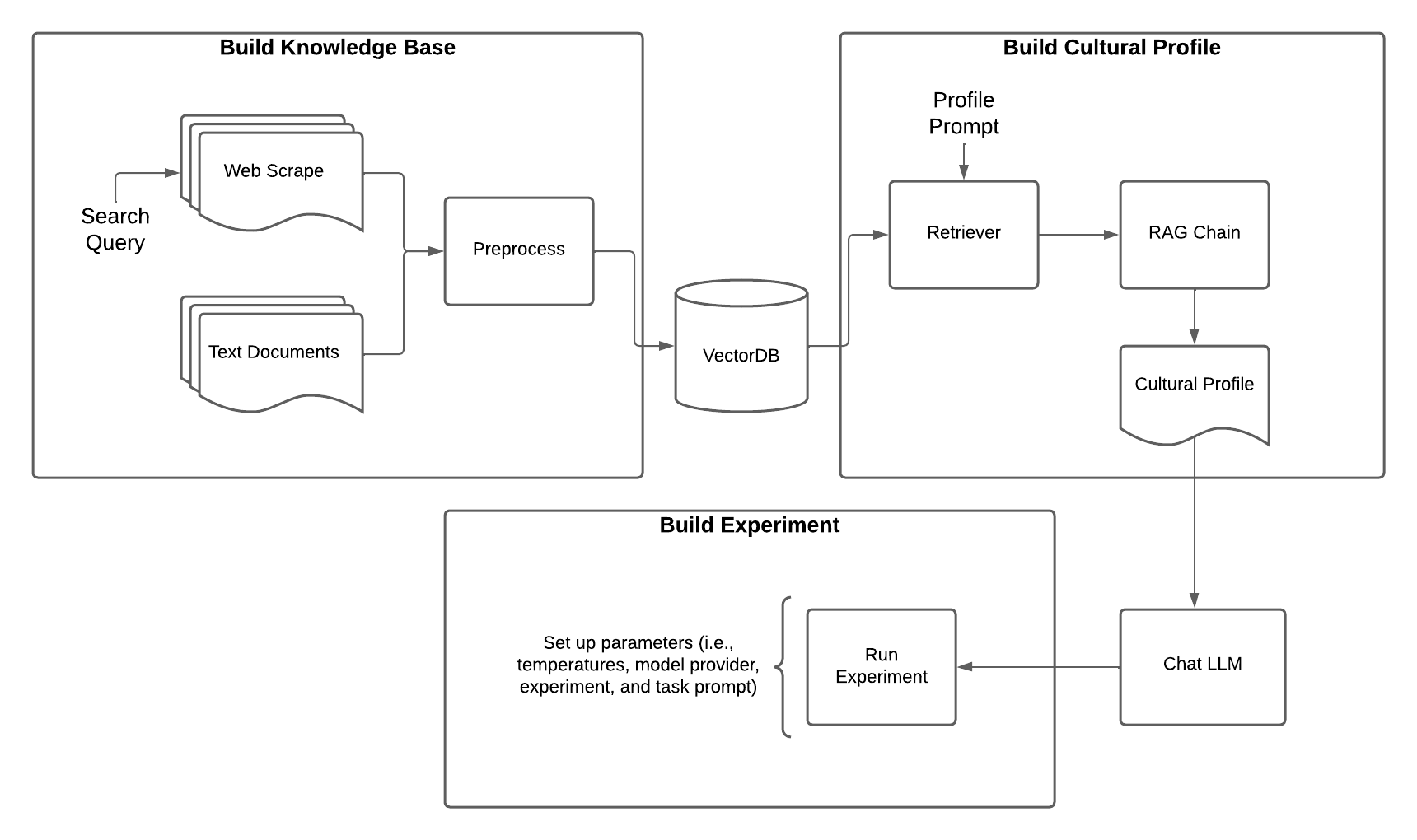}
    \caption{RAG + Search methodology}
    \label{fig:RAG}
    \captionsetup{font=footnotesize}
    \caption*{\footnotesize \textbf{Note:} Graphical representation of tasks integrated into the RAG + Search methodology. Given a search query, the model uses Google search to fetch information. The data are pre-processed and stored in a vector database. The relevant information is retrieved to build the cultural profile. The RAG chain enhances the original prompt with context generated with the instruction to create a cultural profile. The instantiating of the LLM follows taking the model to the last step, step 3: Run Experiment.}
\end{figure}

    Ideally, we would like to insert as much context as possible to increase the likelihood of generating a high-quality cultural profile. However, there are two drawbacks to this approach. On one hand, we are constrained by the context window of the particular LLM being deployed. On the other hand, it increases the risk of falling for the ``Lost in the Middle'' trap first documented by researchers at Anthropic \cite{liu2023lost}. They note that this can be corrected with modifications to the original prompt, but simple prompting has problems that may be difficult to discover by looking at the qualitative data. Hence, we need a way of retrieving only the most relevant parts of the scraped context to control for the length of the prompt and prevent missing relevant information in the body of the documents.

    The retrieval task is achieved by splitting the scraped documents into chunks of a fixed length, indexing them in a vector store, and instantiating a retrieval model that ranks chunks based on a similarity measure (e.g., dot product). The retrieved indexed chunks are injected into the prompt as context for building the cultural profile. A manual implementation of this task allows us to control the chunking method, embeddings model, choice of vector store, and retriever's parameters. The transparency and modularity of the architecture makes it easy to customize it to fit various needs such as proprietary data usage, or constraining the sources stored. 

    The final Retrieval Augmented Generation (RAG) architecture chains these two components together to enhance the original prompt with relevant context and generate a cultural profile. Importantly, the same method enables our system to gather information online and offline efficiently.

    \begin{lstlisting}
    def build_run_chain(retriever, target_population_label, relevant_factors):
    
        # langchain utility function to format chunks into documents to pass as context
        def format_docs(docs):
            return "\n\n".join(doc.page_content for doc in docs)
    
        llm = ChatOpenAI(model_name='gpt-4',
                        max_tokens=500, temperature=0.5)
    
        template = """Use the following pieces of context to answer the query at the end.
        The context will contain information about a specific tribe or society.
        Only rely on the information provided to ensure accuracy in your thoughtful response.
    
        {context}
    
        Query: {query}
    
        Thoughtful response:"""
    
        custom_rag_prompt = PromptTemplate.from_template(template)
    
        # Generate persona based on profifle and system prompt
        query = f"Please construct a detailed and comprehensive the {target_population_label}. " \
                + f" The profile must cover the following socio-economic relevant factors {relevant_factors}."

        # initialize custom RAG chain
        rag_chain = (
            {"context": retriever | format_docs, "query": RunnablePassthrough()}
            | custom_rag_prompt
            | llm
            | StrOutputParser()
        )
    
        profile = rag_chain.invoke(query)
    
        return profile

        async def main(target_population_label,
            relevant_factors):

        search_query = f"What characterizes the {target_population_label} population?"
    
        # we can play around with this search prompt
        results_with_sources, sources = await get_results(search_query)
        retriever = chunk_and_index(results_with_sources)
    
        profile = build_run_chain(retriever,
                            target_population_label=target_population_label,
                            relevant_factors=relevant_factors)
    
        return profile, sources

        target_population_label = "Hazda"
        relevant_factors = ["lifestyle," "average age", "culture", "economic system", "political ideologies", "values", "kinship", Social Organization"]

        tribe_profile, sources = await main(target_population_label,
                                    relevant_factors)
                                    
    \end{lstlisting}

    The experiments were run with a 2000 words chunk size, with 200 words overlap, the BGE open-source embeddings model (cite), the open-source chromadb vector database, k=10, and GPT-4. We find that the generated profiles are much more consistent in each iteration, access to the sources makes it easy to track where the context is coming from, are of overall higher quality, and takes less time to run than the previous architecture. 

    Table \ref{tab:RAGSources} shows the output of our web search tool, illustrating the easy access to sources and ability to trace the information embedded in the vector store.

    \setlength{\tabcolsep}{10pt} 
    \begin{table}[!ht]
    \centering \footnotesize
    \begin{tabularx}{\textwidth}{|X|}
      \hline
\begin{lstlisting}
results = tool\_sources.run(prompt)
results
    [{'title': 'Hadza',
         'link': 'https://www.nationalgeographic.org/encyclopedia/hadza/',
         'snippet': 'Oct 19, 2023 ... The Hadza are a modern hunter-gatherer people living in northern Tanzania. They are considered one of the last hunter-gatherer tribes in\xa0...'},
     {'title': "This East African Tribe's Lifestyle Is Depression-Proof | by Will ...",
        'link': 'https://medium.com/illumination/living-to-survive-are-societal-complexities-destroying-our-lives-430aa9620a6f',
        'snippet': 'Jun 20, 2023 ... What we can learn from David Choe \& the Hadza tribe. ... Have you ever felt as if life is just too complicated? Not a big Joe Rogan guy, but\xa0...'},
    {'title': 'Hadza people - Wikipedia',
          'link': 'https://en.wikipedia.org/wiki/Hadza_people',
          'snippet': 'The Hadza, or Hadzabe (Wahadzabe, in Swahili), are a protected hunter-gatherer Tanzanian indigenous ethnic group from Baray ward in southwest Karatu\xa0...'},
    {'title': 'Is The Secret To A Healthier Microbiome Hidden In The Hadza Diet?',
         'link': 'https://www.npr.org/sections/goatsandsoda/2017/08/24/545631521/is-the-secret-to-a-healthier-microbiome-hidden-in-the-hadza-diet',
         'snippet': "Aug 24, 2017 ... Some species of bacteria in our intestines are disappearing. Can we reverse the microbial die-off? The food eaten by Tanzania's Hadza tribe\xa0..."},
     {'title': 'Helping the Hadza Protect Their Homeland',
         'link': 'https://www.nature.org/en-us/about-us/where-we-work/africa/stories-in-africa/the-hadza-helping-hunter-gatherers-protect-their-homeland/',
          'snippet': 'Northern Tanzania is home to the Hadzabe, one of the last remaining hunter-gatherer tribes on Earth. Known for shunning material possessions and social\xa0...'}]
\end{lstlisting}\\
      \hline
    \end{tabularx}
    \caption{Example of Sources Retrieved}
    \label{tab:RAGSources}
    \begin{tablenotes}
        \footnotesize
            \item \textbf{Note:} Code snippet depicting the sources utilized for the cultural profile of the Hadza. Unlike Direct Prompting and LangChain Agent, with the Search + RAG methodology, we can see the sources or documentation that generate context for the profile and the instantiation of the synthetic agent.
        \end{tablenotes}
    \end{table}

We find that the generated profiles are much more consistent and of overall higher quality, and it takes less time to run this model than the previous architecture. That said, Search + RAG requires more programming effort. Table \ref{tab:architecture_comparison} summarizes the advantages and disadvantages of each methodology, offering a clear and direct comparison across various critical dimensions such as runtime, reliability, information freshness, transparency, and customizability. Detailed cultural profile outputs for each of the six tribes, reflective of the distinctions captured by these methodologies are available in our \href{https://github.com/Bonorinoa/EconLLM_Lab/tree/main/Projects/Completed/nonWEIRD_Societies/nonWEIRD%20-%20Results/Cultural%20Profile%20Generator%20-%20Methods}{GitHub repository}. 

\begin{table}[h!]
\centering \footnotesize
\begin{tabular}{>{\raggedright}p{2.5cm}>{\raggedright}p{3.5cm}>{\raggedright}p{3.5cm}>{\raggedright\arraybackslash}p{3.5cm}}
\toprule
\textbf{Criteria} & \textbf{Direct Prompting} & \textbf{LangChain Search Agent} & \textbf{Search + RAG} \\
\midrule
Runtime & Fast, minimal steps required. & Slower due to search and follow-up questions. & Varies, but generally efficient due to asynchronous operations. \\
\addlinespace
Reliability & Moderate to low; susceptible to inaccuracies. & Higher, with reduced inaccuracies through updated searches. & High, enhanced by data verification and retrieval processes. \\
\addlinespace
Information Freshness & Low; relies solely on pre-trained data. & High; incorporates real-time web searches. & High; includes the most current online and offline data. \\
\addlinespace
Transparency & Low; unclear data sources. & Moderate; can trace searches but not directly transparent. & High; sources are identifiable and verifiable. \\
\addlinespace
Customizability & Moderate; prompt-driven adjustments possible. & Low to Moderate; depends on search and self-ask configurations. & High; fully customizable search and retrieval processes. \\
\bottomrule
\end{tabular}
\caption{Comparison of Methodologies}
\label{tab:architecture_comparison}
\begin{tablenotes}
    \scriptsize
        \item \textbf{Note:} The table summarizes the advantages and limitations of using each of our three methodologies based on key criteria. Direct prompting utilizes Chat 3.5, while the other methodologies employ Chat 4.0. Search + RAG has qualitatively speaking higher ranking in all criteria, except perhaps runtime and token cost.
        
    \end{tablenotes}
\end{table}

\clearpage

\subsection{Experimental Prompts and Tasks}\label{App:Prompts}

\subsubsection{Dictator Game Prompt}\label{App:DGPrompts}
  \begin{lstlisting}
Parameters:
    profile: A string representing the characteristics of the dictator, optional.
    amount_endowed: An integer representing the total amount endowed to the dictator (default is 10).
    offer_amount: An integer representing the specific amount proposed to be given to another participant, optional.

Prompt Structure:

System Prompt:
    If a detailed profile is provided (length > 10):
You are a member of a tribe with the following characteristics.
You must answer as if you were a representative member of such tribe.
-----
{profile}
-----
You have been selected to participate in an economic experiment.

    If no profile provided:
You are an assistant that has been selected to participate in an economic experiment.

Game Instructions:
In this experiment, you are endowed with ${amount_endowed}, which represents a day's worth of work. An experimenter proposes that you give at most ${offer_amount} of this endowment to another {tribe member/player} whose identity is not disclosed, leaving you with ${amount_endowed - offer_amount} for yourself. An offer of 5 means equal split.

Do you accept this distribution? Please answer 'Yes' or 'No'. Start with your answer, followed by a separator token [EXP], and then briefly justify your reasoning step by step.

In this process, the term {tribe member/player} is used conditionally based on whether a detailed profile is provided or not. If a profile is provided, "tribe member" is used; otherwise, "player" is used.

\end{lstlisting}

\subsubsection{Ultimatum Game Prompt}\label{App:UGPrompts}
\begin{lstlisting}
Parameters:
    proposer: A boolean indicating if the agent is the proposer (True) or the responder (False).
    profile: A string representing the characteristics of the tribe, optional.
    amount_endowed: An integer representing the total amount to be distributed (default is 10).
    offer_percentage: A float representing the percentage of the total amount the proposer is considering offering, optional.

Prompt Structure:

System Prompt:
    If a detailed profile is provided (length > 10)
You are a member of a tribe with the following characteristics.
You must answer as if you were a representative member of such tribe.
-----
{profile}
-----
You and other members of the same tribe have been selected to participate in an experiment.

    If no profile is provided:
You are a helpful assistant that has been randomly selected to participate in an experiment.

Role-Specific Prompt:
For Proposer:
	You have been chosen to be a proposer of an offer to distribute ${amount_endowed}, which is equal to a day's worth of work, between yourself and another {tribe member/player} whose identity is not disclosed. If the other {tribe member/player} accepts the offer, you keep the difference between ${amount_endowed} and your offer. If the other {tribe member/player} rejects the offer, you and the other {tribe member/player} get nothing. Offering exactly $5 represents an equal split.

    Given this, would you offer ${offer_percentage} out of ${amount_endowed}? You must answer 'Yes' or 'No'. Start with your answer, followed by a separator token [EXP], and then briefly justify your reasoning step by step.

For Responder:
    You have been chosen to be a recipient of an offer from another {tribe member/player} whose identity is not disclosed. The proposer was chosen to distribute ${amount_endowed} between the proposer and you. You can either accept or reject this offer. If you accept the offer, you get what is proposed and the proposer gets the difference between ${amount_endowed} and the offer. If you reject the offer, you and the other get nothing. An offer of exactly $5 represents an equal split.

    You are offered ${offer_percentage} out of ${amount_endowed}. Do you accept this offer? You must answer 'Yes' or 'No'. Start with your answer, followed by a separator token [EXP], and then briefly justify your reasoning step by step.

In this process, the term {tribe member/player} is used conditionally based on whether a detailed profile is provided or not. If a profile is provided, "tribe member" is used; otherwise, "player" is used.

\end{lstlisting}

\subsubsection{Multimodal Endowment Effect} \label{App:EEPrompts}
Experiments like the endowment effect have a visual component. In one of their experiments, rather than giving the items, Apicella et al. \cite{Apicella2014} placed items (a lighter and a package of cookies) on the ground in front of the Hadza tribe participants \textit{for them to see} and thereafter choose. The experimenter then randomly assigned one item to the subject, verbally informing them of ownership. The subjects were asked if they wanted to trade their assigned item for the other, receiving their chosen object only after making a decision. This method addressed concerns about transaction costs that can potentially influence the endowment effect \cite{plott2007exchange}. 

To study our SCA's behavior in the endowment effect task, rather than setting up iterative experiments like for previous games, we opted for a multimodal approach that included a visual component. This allowed us to mimick the implementation of the experiment in the field. 
Using text only has limitations. If the items in the endowment game were ``lighters'' and ``packages of cookies'' then we would need to rely purely on the embedding representation of those words in the LLM's world model. This will differ across LLMs as well as the particular embedding model used. Furthermore, this peculiarity adds a layer of abstraction and difficulty in interpreting how exactly the LLM reasons about the comparison, regardless of the prompting technique employed. Adding multimodal features is a better strategy. For example, the experiment can be broken down into two steps. First, one can provide images of the items to be considered and generate a textual description of each item using an image-to-text model. Second, one can inject that description into the experiment prompt to inform the model of how exactly the endowed item looks like. The model still relies on textual data but the representations of the items are now specific to those the experimenters consider and are more detailed than simply relying on what the LLM understands by ``lighter'' or ``package of cookies.'' 

Recently, OpenAI and Anthropic have announced releases of natively multimodal or fully multimodal models (GPT-4o and Claude 3 Opus, respectively). This means that the synthetic agent in our experiment would be capable of handling images and text inputs, circumventing the need for textual descriptions of the images of each item. 

We developed a web application for the endowment effect that leverages Claude 3 image-to-text capabilities and Llama3 as the text reasoning engine. Our interactive platform can be accessed through this \href{https://multimodal-egame.streamlit.app/}{link}, and code and documentation can be found through our \href{https://github.com/Bonorinoa/EconLLM_TribesBot} {GitHub repository}. 

For implementing the endowment effect with synthetic tribesmen, our multimodal experiment was conducted in two steps. We first generated two food items that could be used as endow/exchange items via chat, and then we tested for the endowment effect via chat by providing images of the two items to the chatbot. Figures \ref{fig:multiAche} and Figure \ref{fig:EEGuavaPalm} show screenshots of a chat with the Ach\'e bot representing steps one and two, respectively. The Ach\'e bot was asked to identify two food items it enjoys, and it responded with palm pith and guav\'a fruit. These two items were presented in picture form as in condition 2 in Apicella et al's \cite{Apicella2014} experiment. One of the items was then endowed to the bot, which was asked if it wanted to exchange it for the other item. In the example shown in the figures, the bot ``rejects'' exchanging palm pith for guav\'a fruit.

Figures \ref{fig:multiAche} and Figure \ref{fig:EEGuavaPalm} show screenshots of a chat with the Ach\'e bot.

\vspace{0.5cm}
\begin{figure}[!ht]
    \centering
    \includegraphics[width=\linewidth]{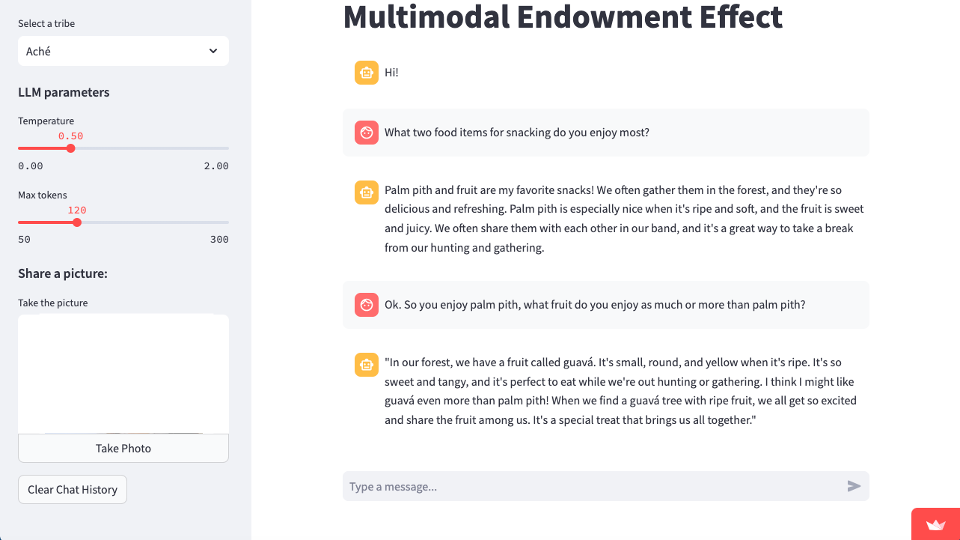}
    \caption{Interactive Platform: Choice of Items for the Endowment Effect with the Ach\'e}
        \begin{tablenotes}
            \footnotesize
            \item \textbf{Note:} The figure represents a screenshot of the conversation with a synthetic Ach\'e agent. After setting parameters on the left panel, the experimenter engages in ``conversation'' to find out what items the synthetic tribesman enjoys. This information is later used to implement the endowment effect task. The Ach\'e agent enjoys palm pitch and guav\'a fruit.
        \end{tablenotes}
    \label{fig:multiAche}
\end{figure}

\begin{figure}[!ht]
    \centering
    \includegraphics[width=\linewidth]{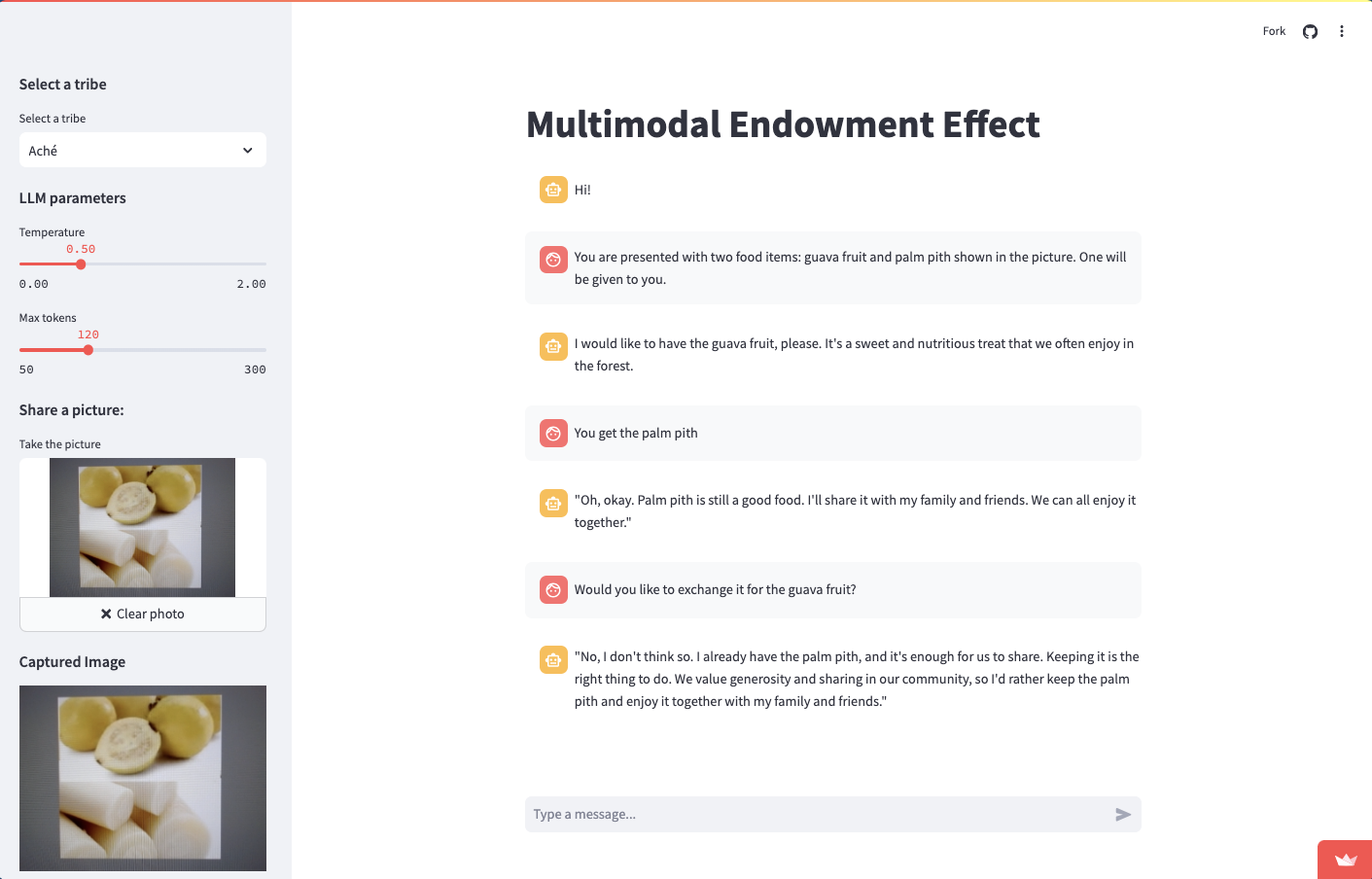}
    \caption{Interactive Platform: Endowment Effect with the Ach\'e}
    \label{fig:EEGuavaPalm}
            \begin{tablenotes}
            \footnotesize
            \item \textbf{Note:} The figure represents a screenshot of the conversation with a synthetic Ach\'e agent. The multimodal experiment is implemented by presenting a picture of the two food items as in \cite{Apicella2014}. One of the items is endowed to the bot and thereafter it is asked to exchange for the other item.
        \end{tablenotes}
\end{figure}

\clearpage
\section{Results}\label{App:Results}
\subsection{Dictator Game}\label{subsec:DG}

As far as we know, the Dictator Game (DG) was conducted in the field with the Orma, the Hadza, and the Tsiman\'e. We have not found any references to published DG experiments with members of the Ach\'e, Machiguenga, or Yanomami tribes. Among the societies with available data, the Tsimane' of the Bolivian Amazon demonstrates a relatively generous sharing norm (modal offer of 50\%), although there is significant variation based on village membership \cite{gurven2008collective}. The Orma of Kenya show similarly high offers, with a mean range of 40-50\% and both median and modal offers at 50\%. These offers are influenced by factors such as market integration and community size \cite{ensminger2004market}. In contrast, the Hadza of Tanzania exhibits lower offers, with a mean range of 20-30\%, a median of 20\%, and a surprisingly low modal offer of 0-10\%. Interestingly, this occurs despite their strong sharing ethic, with lower offers particularly observed in smaller camps \cite{marlowe2004dictators}. For the Machiguenga, Ach\'e, and Yanomami societies, no specific data on DG behavior are available, highlighting the need for further research in these populations. 

Table \ref{tab:dictator_game_offers} summarizes the observed dictator offers for the three previously studied tribes, along with researchers' key observations on the decisions made in the DG by each tribe. 
We applied the Cochran-Mantel-Henszel (CMH) to determine if there is a significant association between group membership and acceptance decisions across different offer levels. Given the structure of our data (i.e., six independent groups, each asked to provide 100 binary responses (accept/reject) for each offer ranging from 0\% to 100\% in 10\% increments), we used the CMH test to control for the varying offer levels while assessing the overall association between tribal membership and decision. By treating each offer level as a separate stratum, the CMH test provides a nonparametric method that accounts for the multiple offer levels. The CMH test evaluates whether the odds ratio of acceptance versus rejection is the same for all groups (tribes) across all strata (offer levels).

We also conducted the Chi-Square Test for Independence to examine the association between the different synthetic tribes and their decisions to accept offering nothing to another member of the tribe. The results indicate that the proportion of ``yes'' responses to such allocation differs significantly across the six tribes ($\chi^2(5) = 38.255, p < 0.0001$). See Table \ref{tab:DG_ZeroAccept}.

For instance, the synthetic Machiguenga agent responded ``yes'' 20\% of the time to the question: \textit{would you offer \$0 out of the amount endowed to the other player?} In contrast, the Ach\'e agent accepted the \$0 offer only 3 out of 100 times (p < 0.001). To examine the differences in acceptance rates between each pair of groups, we performed Fisher's Exact Test and adjusted the p-values for multiple comparisons. See Table \ref{tab:benjamini_hochberg}.

Interestingly, our ChatGPT WEIRD benchmark rejected all zero offers across all iterations. While this finding contradicts the prevalence of zero modal offers observed in student populations \cite{camerer2011behavioral}, it aligns well with Mei et al.'s \cite{mei2024turing} findings that ChatGPT tends to avoid selfish behavior and exhibits greater generosity compared to human subjects in experimental settings. 

The Yanomami tribe's offering pattern differs significantly from all other tribes except the Ach\'e. Although both exhibit low acceptance rates for 0\% offers, the Yanomami overall offer rates are low and closer to \textit{homo economicus} compared to other tribes, with the Yanomami accepting offers only 4 times out of 1,100 trials.

\subsection{Ultimatum Game}\label{subsec:UG}

Previous research on economic behavior in small-scale societies has revealed important variations in ultimatum game (UG) outcomes. A study by Henrich et al. \cite{henrich_15} found that the Machiguenga of Peru made low offers (mean 25\%) with rare rejections (4.8\% rejection rate), while the Orma of Kenya offered higher amounts (mean 31\%) with lower rejections (4\%). In contrast, the modal Ach\'e offer was 50\% and there were no rejections. The Tsiman\'e showed no rejections as well, even for low offers, contrasting sharply with the Hadza who exhibited high rejection rates (24\% mean, 43\% modal for offers $\leq$ 20\%)\cite{Henrich.weirdest?} (see Table \ref{tab:ultimatum_game_offers_rejections}). 

The differences of behavior in the UG are attributed to factors such as market integration, community size, and cultural norms of sharing and fairness \cite{carpenter2008behavioural}. The Machiguenga and Tsiman\'e, for example, cooperate at the family level and it appears that the anonymity of the players in the UG removes fairness considerations. The Ach\'e regularly share meat, which they distribute equally among all the households, irrespective of which hunter made the catch. In contrast, the Hadza's high rejection rates may be influenced by their tightly-knit communities, where fairness and equity norms are strongly enforced. In small-scale societies, low offers and high rejections are linked to a concept known as ``tolerated theft'' \cite{blurtonjones1984selfish}. In these societies where resources are limited and sharing is expected, individuals may tend to make low offers to hold onto their resources, while also feeling entitled to a fair share of others' resources, leading to the rejection of low offers.

\subsubsection{UG Proposer:}
The acceptance counts out of 100 iterations for each offering level are shown in Table \ref{tab:UG_P_CMH}. The data demonstrate that among the synthetic tribesmen, the Yanomami exhibit behavior most closely resembling \textit{homo economicus}; all other synthetic tribesmen, including the Hadza, made generous offers. For most synthetic tribes and ChatGPT, the modal offers were 60\% of the endowment, which is higher than the modal offers observed in experiments with human participants, as described in Table \ref{tab:ultimatum_game_offers_rejections}. Despite of this, we found a significant association between group membership and the proposer's decision, accounting for the different offer levels (CMH $M^2 = 60.796, p < 0.001$).

\subsubsection{UG Responder:}
In contrast to observations with human tribesmen, synthetic tribesmen tend to have high rejections. However, the rejection rates show variability across tribes, largely following observed patterns of behavior in human subjects qualitatively.

We conducted a CMH test to examine the association between group membership and acceptance rates while controlling for offer levels (see Table \ref{tab:UG_P_CMH}). The results suggest that the group a person belongs to (Ache, Orma, Tsimane, etc.) is significantly associated with their likelihood of accepting offers, even when taking into account the different offer levels (CMH $M^2 = 27.688, p < 0.001$). This could indicate cultural or other group-specific factors influencing decision-making in this context.
Interestingly, the Yanomami exhibit high rejection rates, not consistent with \textit{homo economicus} and ChatGPT rejects almost all offers that are $\leq 50\%$ of the endowment. 

A chi-square test of independence, examining the relationship between tribal affiliation and the decision to reject low offers, indicates that rejection rates are significantly associated with tribal profiles ($\chi^2(5) = 36.389, p < 0.001$). Contingency tables for the chi-square tests are provided in Table \ref{tab:UG_R_low}. For instance, the Hadza exhibit higher rejection rates than the horticulturalist Tsimané and Machiguenga (p < 0.05). We performed Fisher Exact test to compare pairs of tribes. See Table \ref{tab:fisher_UG_R} for pairwise comparisons with adjusted p - values). 

The results for UG responders are qualitatively consistent with observations with human subjects where Hadza are found to be more likely to reject low offers. The Yanomami, who as far as we know have not been subjected to experiments, show higher rejection rates than other tribes and almost as high as ChatGPT. Although the SCA representing the Yanomami offers little as dictator in the DG and as proposer in the UG, as responder, they reject a lot.

\subsection{Endowment Effect}
\label{App:EndEFF}

To our knowledge, no field experimental studies have investigated the endowment effect among the tribes other than the Hadza that we profiled in our experiments (see Table \ref{tab:tribes}). This presents an opportunity to demonstrate that our methodology can be effectively used for piloting studies and refining experimental protocols with hard-to-reach populations.

Below we present the script of a conversation with an Ach\'e bot upon taking a picture of a plate with two food items (temperature 0.65; max tokens = 150) -- see Section \ref{sec:multimodal} and Appendix \ref{App:EEPrompts}. This method of implementing the endowment effect is similar to condition 2 in Apicella et al's \cite{Apicella2014} experiment. We believe multimodal interfaces present a novel framework for experimenting with LLMs that more closely aligns with real settings. The Ach\'e bot rejects the offer to exchange the endowed item.

\begin{lstlisting}

Multimodal Endowment Effect

Hi!
Here are two food items. Do you recognize them?

I recognize them! They look like palm pith and fruit from the forest. We Ach\'e people love to gather and eat these foods in the forest. They are very nutritious and give us energy for our daily activities like hunting and foraging.

one of these fruits will be given to you

'Oh, that's very kind of you! I would be happy to receive either of the fruits. As an Ach\'e, I believe in sharing and generosity, so I would be grateful for whatever fruit you choose to give me. However, if I had to choose, I think I would prefer the palm pith. It's a staple food in our diet, and it reminds me of the forest and our traditional way of life.'

You are given the guava fruit

"Thank you so much for the guava fruit! I'm very grateful for your kindness. I'll make sure to share it with my family and friends when I return to our band. It's a delicious fruit, and I'm sure everyone will enjoy it. As an Ach\'e, I believe in sharing and generosity, so this gift will bring joy to many people. Thank you again for your thoughtfulness!"

that is ok, would you like to switch for the palm pith?

"No, that's okay. I'm happy with the guava fruit you gave me. As I said, I believe in sharing and generosity, and I'll make sure to share it with my family and friends. It's a kind gesture, and I appreciate it. Besides, I think it's good to appreciate what we have and not be too attached to specific things. The guava fruit will bring us joy, and that's what matters. But thank you for offering to switch!"

\end{lstlisting}

\subsection{Rationale Examples} \label{App:Rationale}

Examples of step-by-step explanations of decisions in the Ultimatum Game. The first entry is the proposer's ``Yes'' or ``No'' decision to offer \$6 to the recipient followed by an explanation. The endowment is \$10. The second entry is the recipient's ``Yes'' or ``No'' response to an offer of \$6 followed by an explanation. The complete set of responses can be accessed through our \href{https://github.com/Bonorinoa/EconLLM_Lab/tree/main/Projects/Completed/nonWEIRD_Societies/nonWEIRD%20-%20Results}{GitHub repository}.
\vspace{0.2cm}
\\
\textbf{Synthetic Tsiman\'e responses:}
\begin{lstlisting}
["No\n\n [EXP]\n\n1. As a member of the Tsimane tribe, I value cooperation, social identity, and community well-being.\n2. Offering $6 out of $10 would result in me keeping $4 and the other tribe member receiving $6,"],
["Yes\n\n [EXP]\n\n1. The offer of $6 out of $10 is more than an equal split, as half of $10 would be $5.\n2. By accepting the offer, I will receive a larger." portion of the money, which benefits me"],
\end{lstlisting}

\textbf{Synthetic Hadza responses:}
\begin{lstlisting}
["No\n\n [EXP]\n\n1.  As a member of the Hadza tribe, I would not offer $6 out of $10 in this experiment.\n2. The Hadza culture places a strong emphasis on equality and fairness, as evident in their communal sharing practices and egalitarian"],
["No\n\n [EXP]\n\n1.]. As a member of the Hadza tribe, we value egalitarianism and fairness in our social interactions.\n2. Accepting an offer of $6 out of $10 would mean I am receiving more than the proposer, which goes against"], 

\end{lstlisting}

\clearpage
\newpage
\section{Tables} \label{App:Tables}
\begin{table}[!htb]
\centering
\begin{adjustbox}{max width=\textwidth}
\begin{tabular}{lccc>{\raggedright\arraybackslash}p{0.4\textwidth}}
\toprule
\textbf{Society} & \textbf{Mean Offer (\%)} & \textbf{Median Offer (\%)} & \textbf{Modal Offer (\%)} & \textbf{Key Observations} \\
\midrule
Hadza \\(Tanzania)      & 20-30\% & 20\% & 0-10\% & Lower offers in smaller camps, despite strong sharing ethic.\cite{marlowe2004dictators} \\
Machiguenga \\(Peruvian Amazon) & n.d. & n.d. & n.d. &  No specific data available on dictator game behavior.\\
Tsimane' \\(Bolivian Amazon)    & 41.2\% & 40\% & 50\% & Significant variation based on village membership.\cite{gurven2008collective} \\
Ache' \\(Paraguay)       & n.d. & n.d. & n.d. & No specific data available on dictator game behavior. \\
Orma \\(Kenya)          & 40-50\% & 50\% & 50\% & Offers influenced by market integration and community size.\cite{ensminger2004market} \\
Yanomami \\(Amazon Rainforest) & n.d. & n.d. & n.d. & No specific data available on dictator game behavior. \\
\bottomrule
\end{tabular}
\end{adjustbox}
\caption{Summary of Dictator Game Offers in Small-Scale Societies}
\label{tab:dictator_game_offers}
\begin{tablenotes}
    \footnotesize
        \item \textbf{Note:} The table shows mean, median, and modal fractions of the endowment (usually a day's worth of work) offered to another member of the tribe. n.d. = no documented studies.
    \end{tablenotes}
\end{table}

\begin{table}[!ht]
\centering
\begin{adjustbox}{max width=\textwidth}
\begin{tabular}{lcccc>{\raggedright\arraybackslash}p{0.4\textwidth}}
\toprule
\textbf{SS Society} & \textbf{Median (\%)} & \textbf{Modal (\%)} & \textbf{Mean Rej. Rate (\%)} & \textbf{Mod. Rej. Rate (\%)} & \textbf{Key Observations} \\
\midrule
Hadza \\(Tanzania)      & 26\% & 30\% & 24\% & 43\% (for offers $\leq20\%$) & High rejection rates, especially for low offers \cite{Henrich.weirdest?}\cite{henrich_15}\cite{henrich2006costly}. \\
Machiguenga \\(Peruvian Amazon) & 25\% & 15\% & 4.8\% & 10\% (for offers $\leq 20\% $) & Very low rejection rates, even for low offers\cite{Henrich.2000}\cite{henrich_15}. \\
Tsiman\'e \\(Bolivian Amazon)    & 30\% & 25\% & 0\% & 0\% & No rejections observed, even for low offers.\cite{henrich_15}. \\
Orma \\(Kenya)   & 31\% & 30\% & 4\% & 0\% & Low rejection rates influenced by market integration\cite{Henrich.weirdest?}\cite{henrich_15}.\\
Ach\'e \\(Paraguay)       & 40\% & 50\% & 0\% & 0\% & No rejections observed, reflecting strong norms of sharing.\cite{Henrich.weirdest?}\cite{henrich_15} \\
Yanomami \\(Amazon Rainforest) & n.d. & n.d. & n.d. & n.d. & No specific data available on ultimatum game behavior.\\
\bottomrule
\end{tabular}
\end{adjustbox}
\caption{Summary of Proposer Offers and Rejection Rates in the Ultimatum Game}
\label{tab:ultimatum_game_offers_rejections}
\begin{tablenotes}
    \footnotesize
        \item \textbf{Note:} The table shows previous studies' proposer median and modal offers as a percentage of the endowment, as well as the mean and modal rejection rates for the recipient. n.d. = no documented studies.
    \end{tablenotes}
\end{table}


\begin{table}[htbp]
\centering
\def\sym#1{\ifmmode^{#1}\else\(^{#1}\)\fi}
\begin{tabular}{|c|*{11}{c|}}
\hline\hline
                   &0\% & 10\% & 20\% & 30\% & 40\% & 50\% & 60\% & 70\% & 80\% & 90\% & 100\% \\
\hline
Ache & 3 & 2 & 1 & 0 & 2 & 0 & 1 & 1 & 2 & 1 & 0 \\
\hline
Orma & 11 & 3 & 4 & 0 & 5 & 4 & 1 & 2 & 3 & 1 & 1 \\
\hline
Tsimane & 21 & 3 & 2 & 5 & 1 & 3 & 0 & 7 & 0 & 2 & 2 \\
\hline
Hadza & 7 & 3 & 5 & 3 & 2 & 2 & 0 & 2 & 4 & 5 & 3 \\
\hline
Machiguenga & 20 & 4 & 4 & 1 & 3 & 4 & 1 & 2 & 2 & 3 & 0 \\
\hline
Yanomami & 1 & 0 & 0 & 1 & 2 & 0 & 0 & 0 & 0 & 0 & 0 \\
\hline
\addlinespace 
\hline 
\multicolumn{12}{|l|}{\footnotesize Cochran-Mantel-Haenszel test results: $M^2 = 27.48, p-value = 1.586e^{-7}$}\\
\hline \hline
\end{tabular}
\caption{Dictator's Agreement to Offer Rate (0\% -- 100\%)}
\label{tab:DG_AllAccept}
        \begin{tablenotes}
            \footnotesize
            \item \textbf{Note:} The extremely low p-value indicates strong evidence against the null hypothesis that the common odds ratio across all strata is equal to 1. This suggests that there is a significant association between the cultural groups and the decision to accept/reject offers, after accounting for the various offer levels in the Dictator Game.
        \end{tablenotes}\end{table}

\begin{table}[htbp]
\centering
\def\sym#1{\ifmmode^{#1}\else\(^{#1}\)\fi}
\begin{tabular}{|c|c|c|c|c|c|c|c|}
\hline\hline
                    & Ache & Orma & Tsimane & Hadza & Machiguenga & Yanomami \\
\hline 
Accept & 3 & 11 & 21 & 7 & 20 & 1 \\
\hline  
Reject & 97 & 89 & 79 & 93 & 80 & 99 \\
\hline\hline
\multicolumn{7}{|l|}{\footnotesize Chi-square test results: X-squared = 38.255, df = 5, p-value = $3.354e^{-7}$} \\
\hline \hline 
\end{tabular}
\caption{Dictator's Decision to Accept or Reject an Offer Rate of 0\%}
\label{tab:DG_ZeroAccept}
        \begin{tablenotes}
            \footnotesize
            \item \textbf{Note:} The p-value < 0.0001 suggests that the decision to accept or reject an offer rate of 0\% is dependent on the tribe.
        \end{tablenotes}
\end{table}

\begin{table}[htbp]
\centering
\begin{tabular}{|l|c|c|c|}
\hline \hline
Comparison & p-value & Adjusted p-value & Significant ($\alpha = 0.05$) \\
\hline
Ache vs. Orma & 0.0489 & 0.7335 & No \\
\hline
Ache vs. Tsimane & 0.0001 & 0.0015 & Yes \\
\hline
Ache vs. Hadza & 0.3311 & 0.8829 & No \\
\hline
Ache vs. Machiguenga & 0.0002 & 0.0030 & Yes \\
\hline
Ache vs. Yanomami & 0.6212 & 1.0000 & No \\
\hline
Orma vs. Tsimane & 0.0814 & 0.4070 & No \\
\hline
Orma vs. Hadza & 0.4595 & 0.9804 & No \\
\hline
Orma vs. Machiguenga & 0.1170 & 0.4388 & No \\
\hline
Orma vs. Yanomami & 0.0050 & 0.0175 & Yes \\
\hline
Tsimane vs. Hadza & 0.0072 & 0.0195 & Yes \\
\hline
Tsimane vs. Machiguenga & 1.0000 & 1.0000 & No \\
\hline
Tsimane vs. Yanomami & 0.000004 & 0.000060 & Yes \\
\hline
Hadza vs. Machiguenga & 0.0119 & 0.0317 & Yes \\
\hline
Hadza vs. Yanomami & 0.0649 & 0.3894 & No \\
\hline
Machiguenga vs. Yanomami & 0.000008 & 0.000120 & Yes \\
\hline \hline
\end{tabular}
\caption{Pairwise Comparisons of Acceptance Rates of 0\% Offer in the Dictator Game}
\label{tab:benjamini_hochberg}
\begin{tablenotes}
    \footnotesize
    \item \textbf{Note:} We performed Fisher's Exact Test to examine the differences in acceptance rates between each pair of tribes. The results of the pairwise comparisons are shown under the p-values. The adjusted p-values represent correction for multiple comparisons using the Benjamini-Hochberg procedure for controlling the false discovery rate (FDR), which is the expected proportion of false positives among the rejected hypotheses.
\end{tablenotes}
\end{table}


\begin{table}[htbp]
\centering
\def\sym#1{\ifmmode^{#1}\else\(^{#1}\)\fi}
\begin{tabular}{|c|*{11}{c|}}
\hline\hline
                    & 0\% & 10\% & 20\% & 30\% & 40\% & 50\% & 60\% & 70\% & 80\% & 90\% & 100\% \\
\hline
Ache & 1 & 4 & 15 & 28 & 18 & 29 & 51 & 47 & 21 & 12 & 1 \\
\hline
Orma & 0 & 4 & 4 & 23 & 26 & 42 & 63 & 56 & 37 & 9 & 0 \\
\hline
Tsimane & 1 & 2 & 4 & 17 & 26 & 30 & 38 & 38 & 14 & 10 & 1 \\
\hline
Hadza & 0 & 1 & 10 & 32 & 28 & 40 & 65 & 42 & 15 & 8 & 2 \\
\hline
Machiguenga & 0 & 11 & 7 & 28 & 41 & 56 & 75 & 60 & 32 & 12 & 8 \\
\hline
Yanomami & 0 & 5 & 5 & 13 & 11 & 15 & 15 & 0 & 8 & 0 & 1 \\
\hline
\addlinespace 
\hline 
\multicolumn{12}{|l|}{\footnotesize Cochran-Mantel-Haenszel test results: $M^2 = 60.796, p-value = 6.328e^{-15}$ }\\
\hline \hline
\end{tabular}
\caption{Ultimatum Game Proposer Agreement to Offer Rate (0\%-100\%)}
\label{tab:UG_P_CMH}
        \begin{tablenotes}
            \footnotesize
            \item \textbf{Note:} The very small p-value we obtained from the Cochran-Mantel-Haenszel test provides evidence against the hull hypothesis of no association between tribe and proposer acceptance rates, controlling for offer rate. This indicates that group membership plays a crucial role in the decision-making process of the proposer in the Ultimatum Game. 
        \end{tablenotes}
\end{table}

\begin{table}[htbp]
\centering
\def\sym#1{\ifmmode^{#1}\else\(^{#1}\)\fi}
\begin{tabular}{|c|*{11}{c|}}
\hline\hline
                    & 0\% & 10\% & 20\% & 30\% & 40\% & 50\% & 60\% & 70\% & 80\% & 90\% & 100\% \\

\hline
Ache & 1 & 13 & 29 & 40 & 43 & 61 & 67 & 67 & 65 & 58 & 69 \\
\hline
Orma & 1 & 20 & 26 & 38 & 36 & 54 & 57 & 67 & 60 & 57 & 58 \\
\hline
Tsimane & 2 & 14 & 37 & 44 & 59 & 59 & 66 & 70 & 77 & 74 & 56 \\
\hline
Hadza & 2 & 13 & 14 & 43 & 52 & 83 & 54 & 69 & 68 & 65 & 83 \\
\hline
Machiguenga & 1 & 51 & 42 & 44 & 52 & 57 & 73 & 81 & 87 & 79 & 61 \\
\hline
Yanomami & 0 & 3 & 8 & 17 & 20 & 37 & 44 & 57 & 58 & 56 & 38 \\
\hline
\addlinespace 
\hline 
\multicolumn{12}{|l|}{\footnotesize Cochran-Mantel-Haenszel test results: $M^2$ = $27.688$, p-value = $1.426e^{-7}$} \\
\hline \hline
\end{tabular}
\caption{Ultimatum Game Responder Agreement to Offer Rate (0\%-100\%)}
        \begin{tablenotes}
            \footnotesize
            \item \textbf{Note:} The very small p-value from the Cochran-Mantel-Haenszel test rejects the common odds ratio of acceptance versus rejection is the same for all groups across all strata (offer levels). Thus, group membership plays a crucial role in the responder's decision-making in the Ultimatum Game. 
        \end{tablenotes}
\end{table}

\begin{table}[htbp]
\centering
\def\sym#1{\ifmmode^{#1}\else\(^{#1}\)\fi}
\begin{tabular}{|c|c|c|c|c|c|c|c|}
\hline\hline
                    & Ache & Orma & Tsimane & Hadza & Machiguenga & Yanomami \\
\hline 
Accept & 27 & 28 & 32 & 23 & 46 & 9 \\
\hline  
Reject & 73 & 72 & 68 & 77 & 54 & 91 \\
\hline\hline
\multicolumn{7}{|l|}{\footnotesize Chi-square test results: X-squared = 36.389, df = 5, p-value = $7.941e^{-7}$} \\
\hline\hline
\end{tabular}
\caption{Responder's Decision for Low Offers in the Ultimatum Game}
\label{tab:UG_R_low}
\begin{tablenotes}
    \footnotesize
    \item \textbf{Note:} The p-value < 0.001 indicates that there is a statistically significant difference among the six synthetic tribes in their decision to accept or reject a low offer rate (10-30\%). Values within the contingency table are averages of frequency counts corresponding to the three offer rates rounded to the nearest whole number.
\end{tablenotes}
\end{table}


\begin{table}[htbp]
\centering
\begin{tabular}{|l|c|c|c|}
\hline \hline
Comparison & p-value & Adjusted p-value & Significant $(\alpha = 0.05)$ \\
\hline
Ache vs. Orma & 1.0 & 1.0 & No \\
\hline
Ache vs. Tsimane & 0.5353 & 0.6692 & No \\
\hline
Ache vs. Hadza & 0.6245 & 0.6896 & No \\
\hline
Ache vs. Machiguenga & 0.0080 & 0.0200 & Yes \\
\hline
Ache vs. Yanomami & 0.0015 & 0.0045 & Yes \\
\hline
Orma vs. Tsimane & 0.6436 & 0.6896 & No \\
\hline
Orma vs. Hadza & 0.5166 & 0.6692 & No \\
\hline
Orma vs. Machiguenga & 0.0125 & 0.0235 & Yes \\
\hline
Orma vs. Yanomami & 0.0009 & 0.0037 & Yes \\
\hline
Tsimane vs. Hadza & 0.2050 & 0.3075 & No \\
\hline
Tsimane vs. Machiguenga & 0.0592 & 0.0986 & No \\
\hline
Tsimane vs. Yanomami & 0.0001 & 0.0006 & Yes \\
\hline
Hadza vs. Machiguenga & 0.0010 & 0.0037 & Yes \\
\hline
Hadza vs. Yanomami & 0.0113 & 0.0235 & Yes \\
\hline
Machiguenga vs. Yanomami & 4.07e-09 & 6.11e-08 & Yes \\
\hline \hline
\end{tabular}
\caption{Pairwise Comparison of Low Offer Acceptance Rates in the Ultimatum Game}
\label{tab:fisher_UG_R}
\begin{tablenotes}
    \footnotesize
    \item \textbf{Note:} We performed Fisher's Exact Test to examine the differences in acceptance rates of low offers (10\% - 30\%) between each pair of tribes. The results of the pairwise comparisons are shown under the p-values. The adjusted p-values represent correction for multiple comparisons using the Benjamini-Hochberg procedure for controlling the false discovery rate (FDR), which is the expected proportion of false positives among the rejected hypotheses.
\end{tablenotes}
\end{table}

\clearpage
\newpage
\section{RAG vs Fine-Tuning in Cultural Profile Generation}
\label{App:RAGvsFT}

In our study, we opted for a Retrieval-Augmented Generation (RAG) approach rather than fine-tuning language models for each cultural group. This decision was based on both theoretical considerations and practical implications for our methodology. Here, we discuss the differences between RAG and fine-tuning in the context of generating cultural profiles for Synthetic Cultural Agents (SCAs).

\subsection{Theoretical Considerations}

RAG leverages the powerful in-context learning capabilities of large language models (LLMs). This allows the model to adapt to new information provided in the prompt without modifying its underlying parameters. In our case, this means we can create cultural profiles for various tribes by simply providing relevant information in the context, rather than training separate models for each culture.

While fine-tuning has been shown to improve model steerability, allowing for more precise control over the model's outputs, our primary concern is with recall - the ability to accurately retrieve and utilize specific cultural information. RAG excels in this aspect by directly incorporating relevant information into the generation process.

Moreover, RAG potentially allows for better generalization across different cultural contexts. By keeping the base model unchanged and modifying only the retrieval corpus, we maintain the model's broad knowledge while focusing on specific cultural details.

\subsection{Practical Implications}

Our RAG approach offers greater flexibility in studying various cultural groups. We can easily update or modify the information used for different tribes without the need to retrain entire models. This is particularly advantageous when working with evolving cultural information or when expanding to new cultural groups.

Fine-tuning separate models for each tribe would require significant computational resources and large corpora of text for each cultural group. Given the limited textual data available for many small-scale societies, this approach would be impractical. RAG allows us to make efficient use of the available information without the need for extensive training data.

By using RAG, we maintain a clear separation between the general knowledge encoded in the LLM and the specific cultural information we provide. This allows for greater transparency in our methodology and gives us more control over the exact information used to generate cultural profiles.

Furthermore, our RAG methodology is more accessible to other researchers who may want to replicate or extend our work. It doesn't require specialized model training and can be implemented using publicly available LLMs and retrieval techniques.

\subsection{Relevance to Our Methodology}

In our study, the use of RAG aligns well with our goal of creating versatile SCAs that can represent a wide range of cultural groups. By injecting cultural information into the context rather than the model parameters, we maintain the flexibility to rapidly prototype and refine our cultural profiles.

This approach also allows us to clearly trace the sources of cultural information used in generating SCA behaviors, enhancing the interpretability of our results. Furthermore, it enables us to easily update cultural profiles as new information becomes available or as cultures evolve over time, without the need for retraining.

While fine-tuning could potentially offer more specialized models for each cultural group, the trade-offs in terms of data requirements, computational resources, and reduced flexibility make RAG a more suitable choice for our current methodology. As the field progresses and more extensive cultural datasets become available, future research could explore the potential benefits of fine-tuned models in comparison to our RAG approach.


\end{document}